\documentclass[11pt,a4paper]{article}
\usepackage{xurl}
\usepackage{Resource/Styles/report-style}
\usepackage{rotating}
\usepackage{fontawesome5}
\usepackage{needspace}
\usepackage{flafter}
\usepackage{float}

\usepackage[normalem]{ulem}

\makeatletter
\renewcommand{\fps@figure}{!htbp}
\renewcommand{\fps@table}{!htbp}
\makeatother

\hypersetup{
  pdftitle={InclusionMed: Benchmarking AI for Inclusive Healthcare},
  pdfauthor={The InclusionMed Collaboration}
}

\title{InclusionMed:}
\renewcommand{\reportdescription}{Benchmarking AI for Inclusive Healthcare}
\author{\hyperref[app:authors]{\textcolor{black}{The InclusionMed Collaboration}}}
\affiliations{}
\renewcommand{\reportcontact}{%
  \begin{tabular}{@{}c@{\hspace{0.6em}}l@{\hspace{1em}}l@{}}
    \faEnvelope & \textbf{Email} & \href{mailto:info@inclusionmed.ai}{\textcolor{blue}{\nolinkurl{info@inclusionmed.ai}}}\\
    \faGlobe & \textbf{Website} & \href{https://inclusionmed.ai/}{\textcolor{blue}{\nolinkurl{https://inclusionmed.ai/}}}\\
    \faGithub & \textbf{GitHub} & \href{https://github.com/panuvi/inclusionmed}{\textcolor{blue}{\nolinkurl{https://github.com/panuvi/inclusionmed}}}
  \end{tabular}%
}

\begin{document}
\maketitle

\begin{abstract}
\begingroup
\hyphenpenalty=10000
\emergencystretch=2em
Advancing medical AI for inclusive healthcare requires evaluation across diverse healthcare roles, tasks, and care settings, but benchmark coverage remains uneven. We introduce \textbf{InclusionMed}, an open medical AI evaluation initiative for inclusive healthcare. The \textbf{InclusionMed Atlas} organizes benchmarks across three healthcare roles and three AI capabilities. For the first release, we selected four public benchmarks from 388 candidates through structured audits and independent reproduction. With InclusionMed-DxFiles, our private diagnostic reasoning benchmark, the Atlas contains five benchmarks covering three of its nine role–capability cells. We evaluated frontier large language models on the Atlas and aggregated their results into the \textbf{InclusionMed Index}, with overall scores ranging from 17.5 to 49.1 out of 100. The Index also reports scores by healthcare role and AI capability, showing that a model's rank varied with the role or capability considered. A within-benchmark analysis using task-level tags showed that the ten highest-ranked models in the Index all scored higher on patient communication than on clinical decision-making. Models with similar Index scores also differed substantially in estimated inference cost. These findings highlight the need for evaluation covering more diverse healthcare needs and for context-specific model selection. Through open collaboration and community contributions, InclusionMed seeks to ensure that diverse healthcare needs worldwide shape how medical AI progress is defined and measured.
\par
\endgroup
\end{abstract}

\par
\noindent\begin{minipage}{\linewidth}
  \centering
  \IfFileExists{Resource/Figures/inclusion_atlas.pdf}{%
    \includegraphics[width=\linewidth]{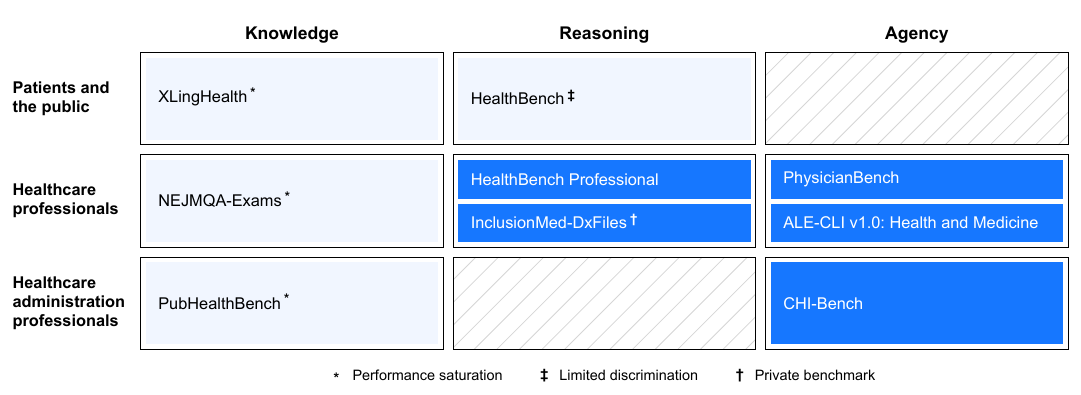}%
  }{%
    \figureplaceholder[18mm]{InclusionMed overview}%
  }
  \captionsetup{font=small}
    \captionof{figure}{%
    \textbf{InclusionMed Atlas.}
    Benchmarks are organized by healthcare role (rows) and AI capability (columns).
    Dark blue tiles mark the four public benchmarks and the private InclusionMed-DxFiles ($\dagger$), each contributing 20\% to the Index.
    Light blue tiles show candidates excluded for score saturation (*) or limited discrimination ($\ddagger$) in this evaluation.
    Hatched cells indicate no eligible benchmark for independent reproduction in this release.
    }
  \label{fig:atlas}
\end{minipage}

\clearpage
\setcounter{tocdepth}{2}
{\hypersetup{linkcolor=black}\small\setlength{\parskip}{0pt}\linespread{0.9}\selectfont\tableofcontents}
\clearpage

\section{Introduction}
\label{sec:introduction}

The uneven distribution of healthcare resources continues to limit access to high-quality care~\citep{whoworldbank2025uhc}.
Developments in medical AI and generalist foundation models~\citep{rajpurkar2022ai,moor2023generalist} offer new ways to support personal health management~\citep{heydari2026personalhealthagent}, clinical practice~\citep{goh2025gpt4assistance}, and the organization of care~\citep{tao2026caretransitions}.
Evaluating these applications requires benchmarks that reflect diverse healthcare roles, tasks, and care settings~\citep{bedi2025evaluation}.
Existing benchmarks span examination questions in MedQA and MedMCQA~\citep{jin2021medqa,pal2022medmcqa,singhal2023clinicalknowledge}, open-ended health conversations in HealthBench~\citep{arora2025healthbench}, and simulated healthcare workflows in MedAgentBench and CHI-Bench~\citep{jiang2025medagentbench,chen2026chibench}.
HELM and MedHELM support comparisons across tasks and datasets~\citep{liang2023helm,bedi2026medhelm}, while the Artificial Analysis Healthcare \& Medical Index aggregates selected benchmark results into an overall score~\citep{artificialanalysisHealthcareMedicalIndex}.
However, the meaning of an overall score depends on what the included benchmarks assess and whose needs they represent~\citep{jacobs2021measurement}.
For inclusive healthcare, this raises the question of how benchmarks can be selected and organized so that an index reflects diverse needs and supports comparisons relevant to different uses of medical AI~\citep{raji2021benchmark}.

We introduce \textbf{InclusionMed}, an open medical AI evaluation initiative for inclusive healthcare. The \textbf{InclusionMed Atlas} organizes benchmarks along two axes: healthcare role and AI capability. The role axis identifies whose needs or work a task represents, distinguishing patients and the public, healthcare professionals, and healthcare administration professionals. The capability axis identifies the principal AI capability assessed, distinguishing knowledge, reasoning, and agency. We map each benchmark or healthcare-relevant subset to an Atlas cell based on the predominant role–capability combination among its tasks. Benchmark inclusion is guided by six audit criteria: healthcare relevance, trustworthiness, availability and permitted use, complementary coverage, timeliness and data independence, and discriminative capacity (\hyperref[fig:audit-pipeline]{Figure~\ref*{fig:audit-pipeline}b}). The Atlas design makes clear which healthcare needs are represented by the included benchmarks and where coverage remains limited.

For the first release, we collected 388 candidate benchmarks from the published literature, public platforms and repositories, and community contributions.
With large language model (LLM) assistance, we compiled benchmark cards linked to their source materials (\hyperref[fig:audit-pipeline]{Figure~\ref*{fig:audit-pipeline}a}).
Applying the audit criteria yielded a shortlist of eight candidates.
Independent reproduction revealed score saturation in three medical knowledge benchmarks and limited discrimination among models on HealthBench.
We therefore retained four public benchmarks or subsets: HealthBench Professional~\citep{hicks2026healthbenchprofessional}, PhysicianBench~\citep{liu2026physicianbench}, the Health and Medicine subset of ALE-CLI v1.0~\citep{sun2026ale}\footnote{Hereafter referred to as \textit{ALE-CLI v1.0: Health and Medicine}.}, and CHI-Bench~\citep{chen2026chibench}.
We also included InclusionMed-DxFiles, a private benchmark we developed to assess diagnostic reasoning.
These five benchmarks cover three of the nine Atlas cells, reflecting gaps in suitable public benchmarks for evaluating AI across diverse healthcare roles and needs.

We evaluated 22 models on the benchmarks included in the current Atlas and aggregated their results into the \textbf{InclusionMed Index}, giving each benchmark equal weight.
Overall Index scores ranged from 17.5 to 49.1 on a 0--100 scale.
The Index also reports scores by healthcare role and AI capability, revealing differences in both model rankings and score levels.
For example, role-specific Index scores were consistently lower for healthcare administration professionals than for healthcare professionals, while GPT-6 Luna ranked eighth in reasoning but twentieth in agency.
For finer-grained analysis, we used task-level tags to compare the ten highest-ranked models in the Index within individual benchmarks across three attributes: clinical specialty, medical expertise, and healthcare workflow.
On HealthBench Professional, all ten models scored higher on patient communication than on clinical decision-making.
On both HealthBench Professional and PhysicianBench, treatment planning and implementation had lower median scores than clinical assessment and diagnosis.
Estimated inference costs also varied widely among models with similar Index scores, with an approximately 24-fold difference in mean cost per task run in one comparison.
These findings highlight the importance of context-specific model selection that considers performance on relevant healthcare tasks alongside the associated inference cost.

InclusionMed welcomes contributions that help medical AI evaluation reflect diverse healthcare needs. Contributors can share benchmarks, datasets, or tools; propose benchmark ideas; submit evidence-based corrections; and offer expert guidance or partnerships. These contributions can broaden benchmark coverage and make model comparisons more relevant to healthcare practice. Through successive releases of the Atlas and Index, this open collaboration aims to ensure that the diverse healthcare needs of people worldwide guide how progress in medical AI is defined and measured.

\section{The InclusionMed Atlas}
\label{sec:atlas}
The Atlas is structured as a nine-cell matrix. Each cell pairs a healthcare role with an AI capability to describe the primary evaluation focus of the benchmarks assigned to it.

\subsection{Healthcare roles}
\label{sec:atlas-roles}
The three healthcare roles are:
\begin{itemize}
  \item \textbf{Patients and the public}: people seeking health information or support with managing their health and accessing care.
  \item \textbf{Healthcare professionals}: people engaged in clinical care or in medical research and related technical work.
  \item \textbf{Healthcare administration professionals}: people responsible for management and administrative work within healthcare provider or payer organizations.
\end{itemize}

\subsection{AI capabilities}
The capability axis identifies the main behavior evaluated by a task:
\begin{itemize}
  \item \textbf{Knowledge}: providing or applying established information about medicine or healthcare.
  \item \textbf{Reasoning}: interpreting evidence and context to make a judgment or formulate a recommendation.
  \item \textbf{Agency}: performing actions or producing deliverables that meet specified requirements.
\end{itemize}
These capabilities can coexist within a task. For example, language agents can combine reasoning with tool use~\citep{yao2023react,schick2023toolformer}.

\subsection{Benchmark mapping to the Atlas}
\label{sec:atlas-mapping}
A benchmark may contain tasks spanning multiple role--capability combinations. For benchmarks that also cover non-healthcare domains, we first select the healthcare-relevant subset. We then identify the predominant role--capability combination among the included tasks and assign the benchmark or subset to the corresponding Atlas cell. This placement summarizes its main evaluation focus. The benchmark versions and healthcare subsets included in the current Atlas are described in \hyperref[sec:public-portfolio]{Section~\ref*{sec:public-portfolio}} and \hyperref[sec:newbenchmarks]{Section~\ref*{sec:newbenchmarks}}.
\begin{figure}[!t]
  \centering
  \includegraphics[width= \linewidth]{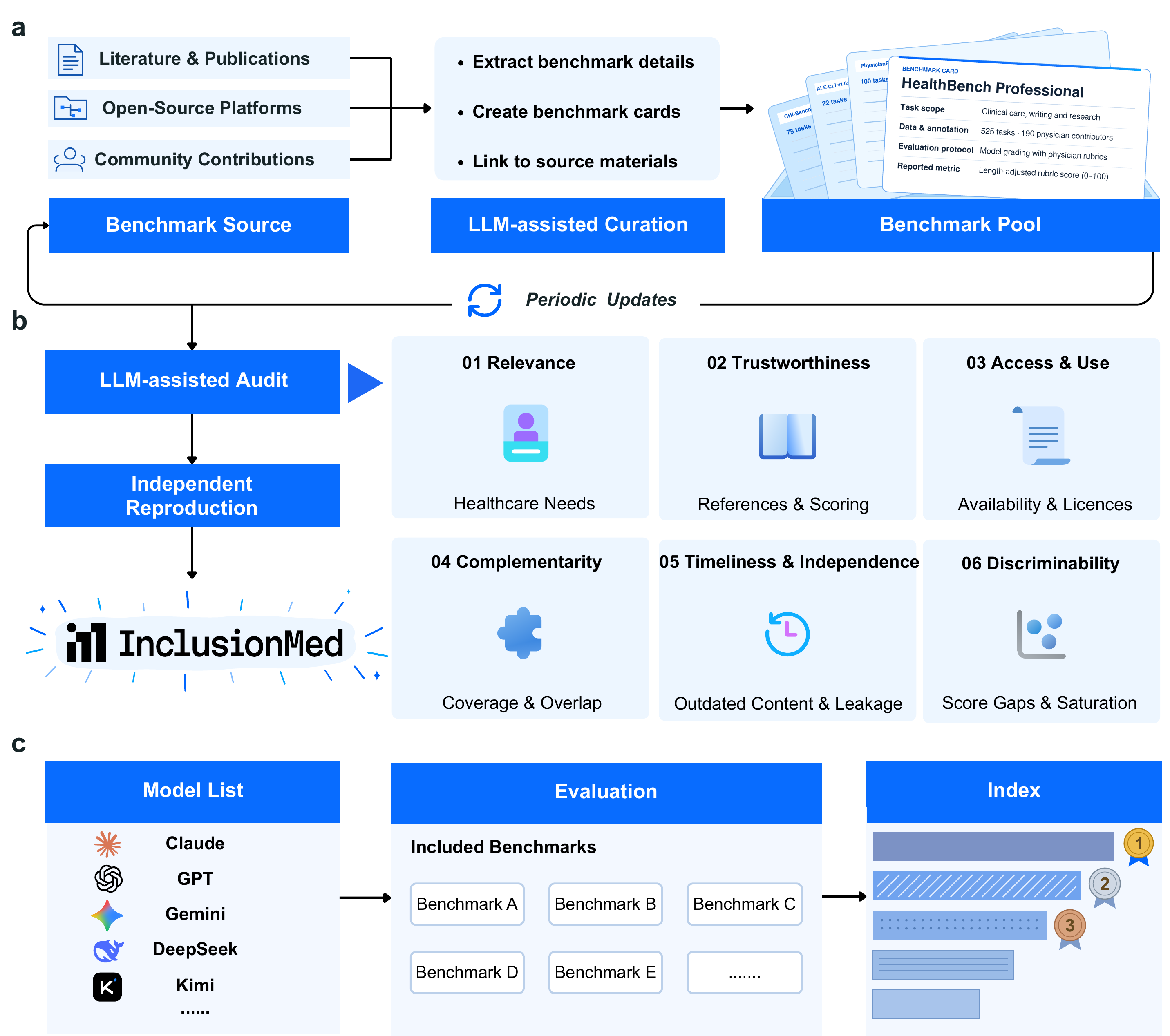}
  \captionsetup{font=small}
    \caption{
    \textbf{Overview of the InclusionMed workflow.}
    \textbf{(a)} Candidate benchmarks are collected from literature, public platforms and community contributions, then organized into benchmark cards through LLM-assisted curation.
    The benchmark pool is updated periodically.
    \textbf{(b)} Candidates undergo sequential screening against six audit criteria, followed by independent reproduction before inclusion.
    \textbf{(c)} Models are evaluated using benchmark-specific execution and scoring procedures, and their scores are aggregated into the InclusionMed Index.
    }
  \label{fig:audit-pipeline}
\end{figure}

\section{Benchmark Portfolio}
\label{sec:scope}
\label{sec:portfolio}

Benchmark inclusion in the InclusionMed Atlas proceeds through collection, auditing and independent reproduction.

\subsection{Benchmark collection}
We collect candidate benchmarks from the published literature, public platforms and repositories, and community contributions (\hyperref[fig:audit-pipeline]{Figure~\ref*{fig:audit-pipeline}a}). Candidates include dedicated healthcare benchmarks and broader benchmarks that may contain healthcare-relevant tasks. LLM-assisted curation extracts and organizes information from these sources into structured benchmark cards, drawing on established approaches to dataset documentation~\citep{gebru2021datasheets}. Each card summarizes task scope, data and annotation, evaluation protocols and reported model performance, with links to source materials. Collection and curation are repeated periodically to incorporate newly identified benchmarks and updates to existing entries.

\subsection{Benchmark auditing}
\label{sec:benchmark-audit}

We audit candidate benchmarks to assess their suitability for medical AI evaluation and their contribution to the InclusionMed benchmark portfolio. The same six criteria apply to public and private benchmarks (\hyperref[fig:audit-pipeline]{Figure~\ref*{fig:audit-pipeline}b}).
The LLM-assisted audit draws on benchmark cards and linked source materials. We developed an agent skill with reusable instructions and supporting resources to guide benchmark-level review and task-level analysis. Human review accompanies the audit throughout to ensure accurate assessments. Further details are provided in \hyperref[app:benchmark-portfolio]{Appendix~\ref*{app:benchmark-portfolio}}.

\paragraph{Healthcare relevance}
We assess whether a benchmark or its healthcare-relevant subset addresses identifiable healthcare needs or activities, based on task objectives and intended users.

\paragraph{Trustworthiness}
We review data provenance, benchmark construction, annotation, and scoring to assess whether reference standards and scoring criteria have a sound medical or professional basis and align with evaluation objectives. For example, we examine the methods used to establish ground truth and exclude benchmarks generated entirely by AI.

\paragraph{Availability and permitted use}
We assess the availability of evaluation resources, including tasks, scoring methods, code, and execution environments. We review documented access routes, restrictions, and licence terms to determine which resources can be used for evaluation.

\paragraph{Complementary coverage}
We assess whether a candidate adds value relative to the existing portfolio through its purpose, task design, or evaluation methods, including broader coverage of healthcare roles, AI capabilities, populations, languages, regions, or resource settings.

\paragraph{Timeliness and data independence}
We assess whether task content and reference standards remain applicable as guidelines, policies, and benchmark versions change~\citep{yan2026livemedbench}. We review documented contamination and leakage safeguards and evidence of test-material exposure~\citep{oren2024contamination}.

\paragraph{Discriminative capacity}
We assess whether benchmark scores distinguish model performance under comparable evaluation conditions. Using publicly reported results from the same benchmark version, we examine score distributions and differences between models. This initial assessment informs which candidate benchmarks proceed to independent reproduction.

\begin{figure}[!t]
  \centering
  \includegraphics[width=\linewidth]{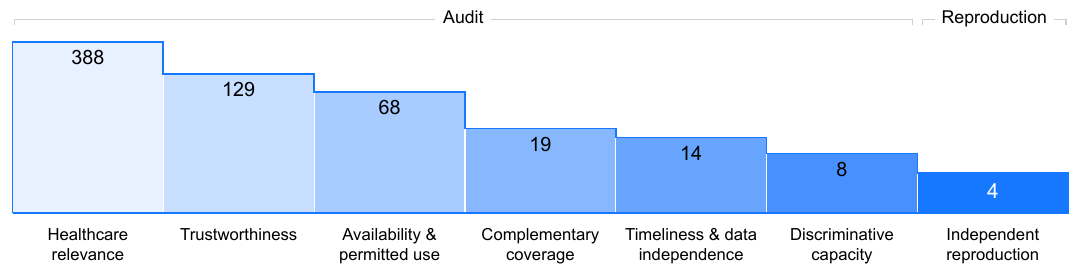}
  \captionsetup{font=small}
    \caption{
    \textbf{Benchmark screening and retention.} Counts show candidates remaining after each audit criterion and independent reproduction. Bar heights use a $\log_{10}$ scale with a baseline of one benchmark.}

  \label{fig:benchmark-selection}
\end{figure}

\subsection{Independent reproduction}
\label{sec:independent-validation}

For this release, we independently reproduced task execution and scoring for eight shortlisted candidates. We followed each benchmark's original evaluation protocol and documented the execution environment, model configurations, and any deviations from the original implementation~\citep{pineau2021reproducibility}. We evaluated the same eight models across all eight shortlisted benchmarks. Benchmarks were excluded for score saturation or limited discrimination using the criteria specified in \hyperref[app:benchmark-portfolio]{Appendix~\ref*{app:benchmark-portfolio}}.

The evaluated versions or subsets of XLingHealth~\citep{jin2023xlinghealth}, NEJMQA-Exams~\citep{katz2024nejmexams}, and PubHealthBench~\citep{harris2025pubhealthbench} met the saturation criterion. HealthBench met the limited-discrimination criterion. These four benchmarks were therefore excluded from the current Atlas. These results yielded the four public
benchmarks described below
(\hyperref[fig:benchmark-selection]{Figure~\ref*{fig:benchmark-selection}}). \hyperref[fig:benchmark-saturation]{Figure~\ref*{fig:benchmark-saturation}} provides the results of these analyses.

To support finer-grained analysis of model performance, we also characterized tasks in the included benchmarks using task-level tags. The InclusionMed tagging scheme comprises eight task attributes grouped into four dimensions (\hyperref[fig:inclusionmed-task-attribute-taxonomy]{Figure~\ref*{fig:inclusionmed-task-attribute-taxonomy}}), with their distributions reported in \hyperref[app:outputs]{Appendix~\ref*{app:outputs}}. Model performance by task attributes will be illustrated in \hyperref[sec:item-tag-insights]{Section~\ref*{sec:item-tag-insights}}.

\subsection{Included public benchmarks}
\label{sec:public-portfolio}

\paragraph{HealthBench Professional}
HealthBench Professional evaluates model responses to clinicians' professional queries \citep{hicks2026healthbenchprofessional}. It contains 525 cases, each requiring the model to generate the next assistant response in a single- or multi-turn clinician--model conversation. Responses are assessed against physician-developed, case-specific rubrics. Response length can influence automated evaluations~\citep{dubois2024length,zheng2023llmjudge}; we report HealthBench Professional\textquotesingle{}s length-adjusted rubric score on a 0--100 scale following its scoring procedure~\citep{hicks2026healthbenchprofessional}. Its primary Atlas assignment is \textit{healthcare professionals $\times$ reasoning}, reflecting its emphasis on interpreting clinical information and formulating professional responses.

\paragraph{PhysicianBench}
PhysicianBench evaluates agents on electronic health record workflows using 100 cases across 21 clinical specialties \citep{liu2026physicianbench}. Agents receive patient-specific clinical instructions and interact with an isolated environment based on the Fast Healthcare Interoperability Resources (FHIR) standard to retrieve information, perform required actions and document their work. Evaluation uses 670 checkpoints combining programmatic verification of record state with rubric-based assessment of reasoning and documentation. We report pass@1 as the mean single-attempt success rate, with a case counted as successful only if all its checkpoints pass \citep{liu2026physicianbench}. Its primary Atlas assignment is \textit{healthcare professionals $\times$ agency}, reflecting its requirement to translate clinical decisions into verifiable actions and workflow outputs.

\paragraph{ALE-CLI v1.0: Health and Medicine}
ALE-CLI v1.0: Health and Medicine comprises 22 cases covering computational tasks in medical research and related technical work \citep{sun2026ale}. Agents receive task instructions, input files and access to relevant software, and must produce verifiable outputs such as executable code or analysis reports. Task-specific graders assess these outputs against reference results and completion criteria, and we report the mean task score on a 0--100 scale \citep{sun2026ale}. Its primary Atlas assignment is \textit{healthcare professionals $\times$ agency}, reflecting its requirement to complete computational tasks and produce outputs that meet specified criteria.

\paragraph{CHI-Bench}
CHI-Bench evaluates agents on 75 healthcare operations cases in the United States, with 25 each in provider prior authorization, payer utilization management and care management \citep{chen2026chibench}. Agents follow an operations handbook and use provided tools to complete multi-stage workflows, update workflow state and produce required records or decisions. A run is successful only if it passes both programmatic workflow checks and rubric-based assessment. We report pass@1 as the mean single-attempt success rate, averaging success outcomes over three independent runs per case and then across cases \citep{chen2026chibench}. Its primary Atlas assignment is \textit{healthcare administration professionals $\times$ agency}, reflecting its focus on completing healthcare administrative and care coordination workflows.

\subsection{Included private benchmark}
\label{sec:development}
\label{sec:newbenchmarks}

\paragraph{InclusionMed-DxFiles}
    InclusionMed-DxFiles assesses diagnostic reasoning on 312 cases adapted from the \emph{Case Records of the Massachusetts General Hospital} in the \emph{New England Journal of Medicine}. Models review each clinical presentation with the final diagnosis withheld and provide a primary diagnosis with supporting reasoning. The score is the percentage of cases graded \emph{perfect}, requiring a correct primary diagnosis and no unsupported diagnostic claims. The benchmark maps to \emph{healthcare professionals $\times$ reasoning} in the Atlas. Construction and scoring details are provided in \hyperref[app:dxfiles]{Appendix~C}.

\section{Index Protocol}
\label{sec:index}

The initial InclusionMed Index is constructed from the four public benchmarks selected in \hyperref[sec:portfolio]{Section~\ref*{sec:portfolio}} and InclusionMed-DxFiles, our private clinical reasoning benchmark. Each model's index score is the equally weighted mean of its five benchmark scores. These scores are used to summarize performance across the healthcare roles and AI capabilities represented in the Atlas.

\subsection{Evaluation protocol}
\label{sec:model-evaluation}

We evaluated 22 model releases from 11 families (\hyperref[tab:index-model-roster]{Table~\ref*{tab:index-model-roster}}). Evaluations used direct application programming interface (API) calls for HealthBench Professional and InclusionMed-DxFiles, an electronic health record agent in a FHIR environment for PhysicianBench, and command-line interface (CLI) agents for CHI-Bench and ALE-CLI v1.0: Health and Medicine. Model names omit date suffixes for readability. The corresponding release designations and evaluation settings are documented in \hyperref[sec:sup_model_list]{Appendix~\ref*{sec:sup_model_list}}, and \hyperref[tab:cli-versions]{Table~\ref*{tab:cli-versions}} lists harness settings.

\begin{table}[!t]
  \raggedright
  \captionsetup{font=small}
  \caption{\textbf{Tasks, scored runs, metrics and Index weights by benchmark.} The table summarizes task counts, scored runs per task, evaluation metrics and Index weights for each of the five benchmarks.}
  \label{tab:index-score-calculation}

  \small
  \setlength{\tabcolsep}{3pt}
  \renewcommand{\arraystretch}{1.3}
  \begin{tabular*}{\linewidth}{@{\extracolsep{\fill}}lcclc@{}}
    \toprule
    Benchmark & Tasks & Runs/task & Metric & Weight \\
    \midrule
    HealthBench Professional & 525 & 8 & Length-adjusted rubric score & 20\% \\
    \addlinespace[4pt]
    PhysicianBench & 100 & 1 & Pass@1 & 20\% \\
    \addlinespace[4pt]
    ALE-CLI v1.0: Health and Medicine & 22 & 1 & Task score (partial credit) & 20\% \\
    \addlinespace[4pt]
    CHI-Bench & 75 & 3 & Pass@1 & 20\% \\
    \addlinespace[4pt]
    InclusionMed-DxFiles & 312 & 1 & Perfect-grade rate & 20\% \\
    \bottomrule
  \end{tabular*}
\end{table}

\subsection{Benchmark scoring}
\label{sec:benchmark-score-calculation}

Each benchmark is scored using its own evaluation procedure (\hyperref[tab:index-score-calculation]{Table~\ref*{tab:index-score-calculation}}). HealthBench Professional and ALE-CLI v1.0: Health and Medicine report mean scores with partial credit. PhysicianBench reports pass@1, requiring all case checkpoints to pass; CHI-Bench reports pass@1, requiring both workflow and rubric checks to pass. InclusionMed-DxFiles reports the percentage of English-language cases graded \emph{perfect}. All five benchmark scores are expressed on a 0--100 scale, with success rates reported as percentages. Runs/task in the table counts scored outcomes retained per task per model, and a replacement run supersedes the earlier outcome.

On HealthBench Professional, the response-length adjustment is applied before the 4,200 grades for each model are averaged, and only the final mean is clipped to the benchmark's reporting range~\citep{hicks2026healthbenchprofessional}. On CHI-Bench, we average the three binary outcomes within each case and then average the resulting case-level rates across all cases; one successful run yields a case-level success rate of one-third. For PhysicianBench and InclusionMed-DxFiles, the final scores include one grade for each of the 100 and 312 cases, respectively. Invalid runs remaining after retries and task timeouts are counted as failures in the score denominators (\hyperref[app:benchmark-scoring]{Appendix~\ref*{app:benchmark-scoring}}).

\subsection{Index aggregation}
\label{sec:index-aggregation}

The score of model $i$ on benchmark $j$ is denoted by $x_{ij} \in [0,100]$. For the $n$ benchmarks listed in \hyperref[tab:index-score-calculation]{Table~\ref*{tab:index-score-calculation}} ($n=5$ in the initial release), the overall InclusionMed Index score is then defined as
\begin{equation}
  I_i = \frac{1}{n}\sum_{j=1}^{n} x_{ij}.
  \label{eq:inclusionmed-index}
\end{equation}

Each benchmark is assigned an equal weight of 20\% in the Index. Scores are aggregated at two decimal places and rounded for display. In addition to the overall Index, we report role- and capability-specific scores, calculated as the mean of the benchmark scores within each category.

\begin{figure}[!t]
  \centering
  \includegraphics[width=\linewidth]{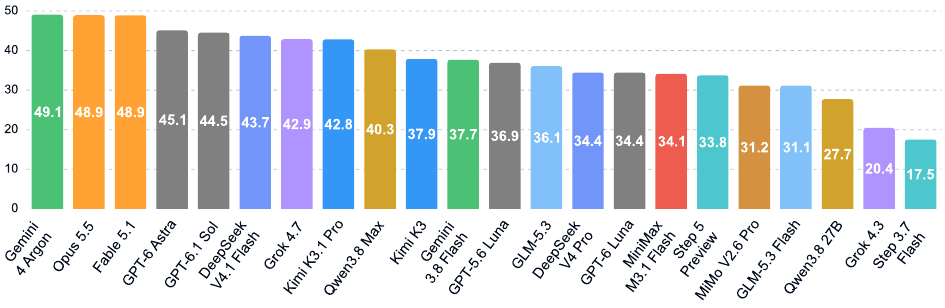}
  \captionsetup{font=small}
    \caption{
    \textbf{InclusionMed Index scores for 22 models.} Scores combine performance across five equally weighted benchmarks. Models are ordered from highest to lowest score in the InclusionMed Index.
    }
  \label{fig:inclusionmed-index-overview}
\end{figure}

\subsection{Inference cost}
\label{sec:inference-cost}

We estimate inference cost from recorded input and output token usage and published model prices, using applicable rates for cached tokens. For each model, mean costs per task run are averaged across all five benchmarks using the Index weights. Costs are reported in US dollars per task run and paired with Index scores to identify the observed Pareto frontier. A model lies on this frontier if no evaluated alternative offers a higher score at the same or lower cost, or a lower cost at the same score~\citep{boyd2004convex}. Pricing and calculation details appear in \hyperref[app:inference-cost]{Appendix~\ref*{app:inference-cost}}.

\section{The InclusionMed Index}
\label{sec:results}

\subsection{Overall performance}
Across the 22 model configurations evaluated on the current InclusionMed Atlas, overall InclusionMed Index scores range from 17.5 to 49.1 on the 0--100 scale (\hyperref[fig:inclusionmed-index-overview]{Figure~\ref*{fig:inclusionmed-index-overview}}). Gemini 4 Argon ranks highest (49.1), followed by Opus 5.5 (48.9) and Fable 5.1 (48.9). GPT-6 Astra (45.1) and GPT-6.1 Sol (44.5) complete the top five.
Note that date suffixes of models are omitted for readability and the specific releases evaluated are documented in \hyperref[app:evaluation-protocol]{Appendix~\ref*{sec:sup_model_list}}.
Complete results for all 22 models across the five benchmarks are available on the \href{https://inclusionmed.ai/}{InclusionMed website}.
 
\subsection{Performance by role and capability}
\label{sec:benchmark-contexts}

Model rankings vary across healthcare roles and AI capabilities (\hyperref[tab:cross-benchmark-atlas-ranks]{Table~\ref*{tab:cross-benchmark-atlas-ranks}}). Although Gemini 4 Argon ranks first overall, Claude Opus 5.5 leads for healthcare professionals (55.6), GPT-6 Astra leads in reasoning (58.7), and Claude Fable 5.1 leads in agency (46.1). These differences are also apparent elsewhere in the rankings. GLM-5.3 ranks 14th for healthcare professionals but 4th for healthcare administration professionals, while GPT-6 Luna ranks 8th in reasoning but 20th in agency. By linking these performance profiles to healthcare roles and capabilities, the Atlas helps users look beyond the overall ranking and follow and compare models on the capabilities relevant to their work.

\begin{table}[!t]
  \centering
  \captionsetup{position=top,width=\linewidth,font=small,justification=raggedright,singlelinecheck=false,skip=7pt}
    \caption{\textbf{Model performance by healthcare role and AI capability.} Scores are on a 0--100 scale, with models ordered by descending InclusionMed Index score. Cell colors share a piecewise-linear scale from 0 to 60. Dashes indicate categories not covered by the current Atlas.}
  \label{tab:cross-benchmark-atlas-ranks}
  \begingroup
  \sffamily\fontsize{8}{9.5}\selectfont
  \definecolor{atlasMuted}{HTML}{696F76}
  \definecolor{atlasScaleLow}{HTML}{FAFCFF}
  \definecolor{atlasScaleHigh}{HTML}{739DC9}
  \color{black}
  \arrayrulecolor{black}
  \setlength{\tabcolsep}{0pt}
  \setlength{\extrarowheight}{1.1pt}
  \renewcommand{\arraystretch}{1.18}
  \newcommand{\AtlasHead}[1]{{\fontsize{7.5}{9}\selectfont\bfseries
    \raisebox{\dimexpr(\depth-\height)/2\relax}{\shortstack[c]{#1}}}}
  \newcommand{\AtlasNA}{\textcolor{atlasMuted}{\textemdash}}
  \newcommand{\AtlasMix}[1]{%
    \fpeval{round(
         4 * min(max((#1),0),12)/12
      + 25 * min(max((#1)-12,0),14)/14
      +      min(max((#1)-26,0),3)/3
      + 17 * min(max((#1)-29,0),11)/11
      +      min(max((#1)-40,0),4)
      + 14 * min(max((#1)-44,0),3)/3
      +  3 * min(max((#1)-47,0),6)/6
      +  5 * min(max((#1)-53,0),3)
      +  2 * min(max((#1)-56,0),2)
      + 11 * min(max((#1)-58,0),1)
      +  2 * min(max((#1)-59,0),1), 2)}}
  \newcommand{\AtlasScore}[1]{%
    \cellcolor{atlasScaleHigh!\AtlasMix{#1}!atlasScaleLow}#1}
  \newcommand{\AtlasSwatch}[1]{%
    \textcolor{atlasScaleHigh!\AtlasMix{#1}!atlasScaleLow}{%
      \rule{0.2485mm}{1.38mm}}}
  \ExplSyntaxOn
  \cs_set_protected:Npn \AtlasScaleStrip
    {
      \int_step_inline:nn {120}
        { \AtlasSwatch{\fpeval{60*(##1-1)/119}} }
    }
  \ExplSyntaxOff
  \newcommand{\AtlasScaleTick}[1]{%
    \rlap{\hspace{\fpeval{29.82*(#1)/60}mm}%
      \makebox[0pt][c]{\raisebox{-1.2pt}{\rule{0.24pt}{1.2pt}}}}%
    \rlap{\hspace{\fpeval{29.82*(#1)/60}mm}%
      \makebox[0pt][c]{\raisebox{-6.5pt}{\fontsize{5.5}{6.5}\selectfont #1}}}}
  \begin{tabularx}{\linewidth}{
    @{}>{\raggedright\arraybackslash}p{0.20\linewidth}
    *{6}{@{\hspace{1.6pt}}>{\centering\arraybackslash}X}@{}}
  \toprule[0.75pt]
  & \multicolumn{3}{@{\hspace{1.6pt}}c}{\textbf{Healthcare role}}
  & \multicolumn{3}{@{\hspace{1.6pt}}c}{\textbf{AI capability}} \\
  \cmidrule[0.30pt](l{4.8pt}r{4.8pt}){2-4}\cmidrule[0.30pt](l{4pt}r{4pt}){5-7}
  \AtlasHead{Model}
    & \AtlasHead{Patients and\\the public}
    & \AtlasHead{Healthcare\\professionals}
    & \AtlasHead{Healthcare\\administration\\professionals}
    & \AtlasHead{Knowledge}
    & \AtlasHead{Reasoning}
    & \AtlasHead{Agency} \\
  \midrule[0.45pt]
  Gemini 4 Argon & \AtlasNA & \AtlasScore{55.0} & \AtlasScore{25.3} & \AtlasNA & \AtlasScore{54.2} & \AtlasScore{45.7} \\
  Claude Opus 5.5 & \AtlasNA & \AtlasScore{55.6} & \AtlasScore{22.2} & \AtlasNA & \AtlasScore{55.4} & \AtlasScore{44.7} \\
  Claude Fable 5.1 & \AtlasNA & \AtlasScore{54.8} & \AtlasScore{25.3} & \AtlasNA & \AtlasScore{53.3} & \AtlasScore{46.1} \\
  GPT-6 Astra & \AtlasNA & \AtlasScore{52.4} & \AtlasScore{16.0} & \AtlasNA & \AtlasScore{58.7} & \AtlasScore{36.1} \\
  GPT-6.1 Sol & \AtlasNA & \AtlasScore{52.1} & \AtlasScore{14.2} & \AtlasNA & \AtlasScore{58.3} & \AtlasScore{35.4} \\
  DeepSeek V4.1 Flash & \AtlasNA & \AtlasScore{49.3} & \AtlasScore{21.3} & \AtlasNA & \AtlasScore{45.4} & \AtlasScore{42.6} \\
  Grok 4.7 & \AtlasNA & \AtlasScore{47.4} & \AtlasScore{24.9} & \AtlasNA & \AtlasScore{51.0} & \AtlasScore{37.5} \\
  Kimi K3.1 Pro & \AtlasNA & \AtlasScore{48.4} & \AtlasScore{20.4} & \AtlasNA & \AtlasScore{49.7} & \AtlasScore{38.2} \\
  Qwen3.8 Max & \AtlasNA & \AtlasScore{47.1} & \AtlasScore{12.9} & \AtlasNA & \AtlasScore{42.7} & \AtlasScore{38.7} \\
  Kimi K3 & \AtlasNA & \AtlasScore{41.7} & \AtlasScore{22.7} & \AtlasNA & \AtlasScore{43.9} & \AtlasScore{33.9} \\
  Gemini 3.8 Flash & \AtlasNA & \AtlasScore{43.7} & \AtlasScore{13.8} & \AtlasNA & \AtlasScore{48.3} & \AtlasScore{30.6} \\
  GPT-5.6 Luna & \AtlasNA & \AtlasScore{43.8} & \AtlasScore{9.3} & \AtlasNA & \AtlasScore{48.9} & \AtlasScore{29.0} \\
  GLM-5.3 & \AtlasNA & \AtlasScore{39.1} & \AtlasScore{24.0} & \AtlasNA & \AtlasScore{41.7} & \AtlasScore{32.3} \\
  DeepSeek V4 Pro & \AtlasNA & \AtlasScore{38.3} & \AtlasScore{19.1} & \AtlasNA & \AtlasScore{39.6} & \AtlasScore{31.0} \\
  GPT-6 Luna & \AtlasNA & \AtlasScore{41.9} & \AtlasScore{4.4} & \AtlasNA & \AtlasScore{48.9} & \AtlasScore{24.7} \\
  MiniMax M3.1 Flash & \AtlasNA & \AtlasScore{37.7} & \AtlasScore{20.0} & \AtlasNA & \AtlasScore{35.9} & \AtlasScore{33.0} \\
  Step 5 Preview & \AtlasNA & \AtlasScore{38.1} & \AtlasScore{16.4} & \AtlasNA & \AtlasScore{38.2} & \AtlasScore{30.8} \\
  MiMo V2.6 Pro & \AtlasNA & \AtlasScore{34.8} & \AtlasScore{16.4} & \AtlasNA & \AtlasScore{31.5} & \AtlasScore{30.9} \\
  GLM-5.3 Flash & \AtlasNA & \AtlasScore{35.3} & \AtlasScore{14.7} & \AtlasNA & \AtlasScore{39.3} & \AtlasScore{25.7} \\
  Qwen3.8 27B & \AtlasNA & \AtlasScore{29.8} & \AtlasScore{19.6} & \AtlasNA & \AtlasScore{30.4} & \AtlasScore{25.9} \\
  Grok 4.3 & \AtlasNA & \AtlasScore{24.1} & \AtlasScore{5.8} & \AtlasNA & \AtlasScore{37.8} & \AtlasScore{8.9} \\
  Step 3.7 Flash & \AtlasNA & \AtlasScore{18.9} & \AtlasScore{12.0} & \AtlasNA & \AtlasScore{23.7} & \AtlasScore{13.4} \\
  \bottomrule[0.75pt]
  \end{tabularx}
  \par\vspace{1.7mm}
  \noindent\makebox[\linewidth][r]{%
    \fontsize{7}{8.2}\selectfont
    Score\enspace 0\enspace
    \AtlasScaleTick{15}\AtlasScaleTick{30}\AtlasScaleTick{45}\AtlasScaleTick{55}%
    \AtlasScaleStrip\enspace 60\strut}
  \par
  \endgroup
\end{table}

\begin{table}[!t]
  \centering
  \captionsetup{position=top,width=\linewidth,font=small,justification=raggedright,singlelinecheck=false,skip=7pt}
    \caption{\textbf{HealthBench Professional performance by clinical specialty and medical expertise.}
    Task attribute scores are shown for the top 10 models by overall InclusionMed Index, with medians calculated across these 10 models.
    Scores span 0--100; colors use a piecewise-linear scale (20--100).}
  \label{tab:index-profiles}
  \label{tab:hbp-medical-expertise}
  \begingroup
  \sffamily\fontsize{7.75}{9.4}\selectfont\color{black}
  \definecolor{hbpLow}{HTML}{FAFCFF}
  \definecolor{hbpHigh}{HTML}{85ADD7}
  \definecolor{hbpSpecialties}{HTML}{E7F0FA}
  \definecolor{hbpExpertise}{HTML}{EEF4FA}
  \arrayrulecolor{black}
  \setlength{\tabcolsep}{1.25pt}
  \renewcommand{\arraystretch}{1.42}
  \newcommand{\hbphead}[1]{{\fontsize{6.5}{7.5}\selectfont\renewcommand{\arraystretch}{0.97}\begin{tabular}[c]{@{}c@{}}#1\end{tabular}}}
  \newcommand{\hbpswatch}[1]{\textcolor{hbpHigh!#1!hbpLow}{\rule{0.62125mm}{1.38mm}}}
  \newcommand{\hbpscaletick}[2]{%
    \rlap{\hspace{#1}\makebox[0pt][c]{%
      \raisebox{-1.2pt}{\rule{0.24pt}{1.2pt}}}}%
    \rlap{\hspace{#1}\makebox[0pt][c]{%
      \raisebox{-6.5pt}{\fontsize{5.5}{6.5}\selectfont #2}}}}
  \resizebox{\linewidth}{!}{%
    \begin{tabular}{@{}l *{7}{>{\columncolor{white}[0.86pt][0.86pt]}c} @{\hspace{4pt}} *{6}{>{\columncolor{white}[0.86pt][0.86pt]}c}@{}}
      & \multicolumn{7}{>{\columncolor{hbpSpecialties}[0.9pt][0pt]}c}{\rule[-3.9pt]{0pt}{14pt}\textbf{Clinical specialty}}
      & \multicolumn{6}{>{\columncolor{hbpExpertise}[0pt][0pt]}c}{\rule[-3.9pt]{0pt}{14pt}\textbf{Medical expertise}} \\
      \addlinespace[3pt]
      \textbf{Model}
      & \hbphead{Internal\\medicine\\specialties}
      & \hbphead{Surgical\\specialties}
      & \hbphead{Obstetrics\\and\\gynecology}
      & \hbphead{Pediatrics}
      & \hbphead{Psychiatry}
      & \hbphead{General\\practice}
      & \hbphead{Other\\specialties}
      & \hbphead{Medical\\understanding}
      & \hbphead{Clinical\\decision-\\making}
      & \hbphead{Safety\\and ethics}
      & \hbphead{Evidence-\\based\\medicine}
      & \hbphead{Patient\\communication}
      & \hbphead{Medical\\calculation} \\
      \specialrule{0.30pt}{3pt}{2.5pt}
      Gemini 4 Argon
        & \cellcolor{hbpHigh!23.52!hbpLow}56.6 & \cellcolor{hbpHigh!14.06!hbpLow}52.3 & \cellcolor{hbpHigh!31.66!hbpLow}60.3 & \cellcolor{hbpHigh!11.42!hbpLow}51.1 & \cellcolor{hbpHigh!16.70!hbpLow}53.5 & \cellcolor{hbpHigh!40.24!hbpLow}64.2 & \cellcolor{hbpHigh!38.04!hbpLow}63.2
        & \cellcolor{hbpHigh!31.88!hbpLow}60.4 & \cellcolor{hbpHigh!15.16!hbpLow}52.8 & \cellcolor{hbpHigh!11.20!hbpLow}51.0 & \cellcolor{hbpHigh!24.40!hbpLow}57.0 & \cellcolor{hbpHigh!40.68!hbpLow}64.4 & \cellcolor{hbpHigh!17.80!hbpLow}54.0 \\
      Claude Opus 5.5
        & \cellcolor{hbpHigh!38.92!hbpLow}63.6 & \cellcolor{hbpHigh!23.08!hbpLow}56.4 & \cellcolor{hbpHigh!46.40!hbpLow}67.0 & \cellcolor{hbpHigh!8.79!hbpLow}49.3 & \cellcolor{hbpHigh!41.12!hbpLow}64.6 & \cellcolor{hbpHigh!27.26!hbpLow}58.3 & \cellcolor{hbpHigh!40.02!hbpLow}64.1
        & \cellcolor{hbpHigh!36.94!hbpLow}62.7 & \cellcolor{hbpHigh!29.90!hbpLow}59.5 & \cellcolor{hbpHigh!28.58!hbpLow}58.9 & \cellcolor{hbpHigh!33.20!hbpLow}61.0 & \cellcolor{hbpHigh!52.56!hbpLow}69.8 & \cellcolor{hbpHigh!8.52!hbpLow}48.4 \\
      Claude Fable 5.1
        & \cellcolor{hbpHigh!32.76!hbpLow}60.8 & \cellcolor{hbpHigh!19.56!hbpLow}54.8 & \cellcolor{hbpHigh!40.46!hbpLow}64.3 & \cellcolor{hbpHigh!14.50!hbpLow}52.5 & \cellcolor{hbpHigh!7.62!hbpLow}45.4 & \cellcolor{hbpHigh!17.80!hbpLow}54.0 & \cellcolor{hbpHigh!34.52!hbpLow}61.6
        & \cellcolor{hbpHigh!32.32!hbpLow}60.6 & \cellcolor{hbpHigh!25.50!hbpLow}57.5 & \cellcolor{hbpHigh!8.52!hbpLow}48.4 & \cellcolor{hbpHigh!37.16!hbpLow}62.8 & \cellcolor{hbpHigh!49.48!hbpLow}68.4 & \cellcolor{hbpHigh!11.42!hbpLow}51.1 \\
      GPT-6 Astra
        & \cellcolor{hbpHigh!42.00!hbpLow}65.0 & \cellcolor{hbpHigh!28.80!hbpLow}59.0 & \cellcolor{hbpHigh!59.27!hbpLow}74.0 & \cellcolor{hbpHigh!23.30!hbpLow}56.5 & \cellcolor{hbpHigh!53.31!hbpLow}70.2 & \cellcolor{hbpHigh!49.70!hbpLow}68.5 & \cellcolor{hbpHigh!51.02!hbpLow}69.1
        & \cellcolor{hbpHigh!39.14!hbpLow}63.7 & \cellcolor{hbpHigh!38.92!hbpLow}63.6 & \cellcolor{hbpHigh!36.50!hbpLow}62.5 & \cellcolor{hbpHigh!36.28!hbpLow}62.4 & \cellcolor{hbpHigh!71.80!hbpLow}82.0 & \cellcolor{hbpHigh!22.86!hbpLow}56.3 \\
      GPT-6.1 Sol
        & \cellcolor{hbpHigh!41.56!hbpLow}64.8 & \cellcolor{hbpHigh!25.50!hbpLow}57.5 & \cellcolor{hbpHigh!57.70!hbpLow}73.0 & \cellcolor{hbpHigh!19.34!hbpLow}54.7 & \cellcolor{hbpHigh!53.00!hbpLow}70.0 & \cellcolor{hbpHigh!46.62!hbpLow}67.1 & \cellcolor{hbpHigh!51.46!hbpLow}69.3
        & \cellcolor{hbpHigh!41.12!hbpLow}64.6 & \cellcolor{hbpHigh!37.38!hbpLow}62.9 & \cellcolor{hbpHigh!30.78!hbpLow}59.9 & \cellcolor{hbpHigh!37.82!hbpLow}63.1 & \cellcolor{hbpHigh!66.32!hbpLow}78.5 & \cellcolor{hbpHigh!12.96!hbpLow}51.8 \\
      DeepSeek V4.1 Flash
        & \cellcolor{hbpHigh!8.85!hbpLow}49.5 & \cellcolor{hbpHigh!8.40!hbpLow}48.0 & \cellcolor{hbpHigh!9.22!hbpLow}50.1 & \cellcolor{hbpHigh!8.70!hbpLow}49.0 & \cellcolor{hbpHigh!3.00!hbpLow}30.0 & \cellcolor{hbpHigh!6.93!hbpLow}43.1 & \cellcolor{hbpHigh!37.82!hbpLow}63.1
        & \cellcolor{hbpHigh!29.90!hbpLow}59.5 & \cellcolor{hbpHigh!11.64!hbpLow}51.2 & \cellcolor{hbpHigh!0.30!hbpLow}21.0 & \cellcolor{hbpHigh!28.14!hbpLow}58.7 & \cellcolor{hbpHigh!34.08!hbpLow}61.4 & \cellcolor{hbpHigh!6.78!hbpLow}42.6 \\
      Grok 4.7
        & \cellcolor{hbpHigh!22.86!hbpLow}56.3 & \cellcolor{hbpHigh!9.00!hbpLow}50.0 & \cellcolor{hbpHigh!32.76!hbpLow}60.8 & \cellcolor{hbpHigh!10.32!hbpLow}50.6 & \cellcolor{hbpHigh!35.62!hbpLow}62.1 & \cellcolor{hbpHigh!15.38!hbpLow}52.9 & \cellcolor{hbpHigh!52.34!hbpLow}69.7
        & \cellcolor{hbpHigh!30.78!hbpLow}59.9 & \cellcolor{hbpHigh!16.70!hbpLow}53.5 & \cellcolor{hbpHigh!8.73!hbpLow}49.1 & \cellcolor{hbpHigh!21.76!hbpLow}55.8 & \cellcolor{hbpHigh!58.01!hbpLow}73.2 & \cellcolor{hbpHigh!7.26!hbpLow}44.2 \\
      Kimi K3.1 Pro
        & \cellcolor{hbpHigh!30.56!hbpLow}59.8 & \cellcolor{hbpHigh!10.54!hbpLow}50.7 & \cellcolor{hbpHigh!30.78!hbpLow}59.9 & \cellcolor{hbpHigh!7.86!hbpLow}46.2 & \cellcolor{hbpHigh!6.99!hbpLow}43.3 & \cellcolor{hbpHigh!9.66!hbpLow}50.3 & \cellcolor{hbpHigh!54.10!hbpLow}70.7
        & \cellcolor{hbpHigh!38.70!hbpLow}63.5 & \cellcolor{hbpHigh!25.50!hbpLow}57.5 & \cellcolor{hbpHigh!4.92!hbpLow}36.4 & \cellcolor{hbpHigh!28.14!hbpLow}58.7 & \cellcolor{hbpHigh!66.00!hbpLow}78.3 & \cellcolor{hbpHigh!7.65!hbpLow}45.5 \\
      Qwen3.8 Max
        & \cellcolor{hbpHigh!6.42!hbpLow}41.4 & \cellcolor{hbpHigh!4.80!hbpLow}36.0 & \cellcolor{hbpHigh!6.99!hbpLow}43.3 & \cellcolor{hbpHigh!5.04!hbpLow}36.8 & \cellcolor{hbpHigh!0.90!hbpLow}23.0 & \cellcolor{hbpHigh!4.38!hbpLow}34.6 & \cellcolor{hbpHigh!10.76!hbpLow}50.8
        & \cellcolor{hbpHigh!8.28!hbpLow}47.6 & \cellcolor{hbpHigh!6.57!hbpLow}41.9 & \cellcolor{hbpHigh!0.99!hbpLow}23.3 & \cellcolor{hbpHigh!5.55!hbpLow}38.5 & \cellcolor{hbpHigh!11.86!hbpLow}51.3 & \cellcolor{hbpHigh!2.28!hbpLow}27.6 \\
      Kimi K3
        & \cellcolor{hbpHigh!12.96!hbpLow}51.8 & \cellcolor{hbpHigh!8.82!hbpLow}49.4 & \cellcolor{hbpHigh!7.68!hbpLow}45.6 & \cellcolor{hbpHigh!6.84!hbpLow}42.8 & \cellcolor{hbpHigh!1.62!hbpLow}25.4 & \cellcolor{hbpHigh!5.55!hbpLow}38.5 & \cellcolor{hbpHigh!38.70!hbpLow}63.5
        & \cellcolor{hbpHigh!26.60!hbpLow}58.0 & \cellcolor{hbpHigh!8.64!hbpLow}48.8 & \cellcolor{hbpHigh!2.28!hbpLow}27.6 & \cellcolor{hbpHigh!25.28!hbpLow}57.4 & \cellcolor{hbpHigh!37.38!hbpLow}62.9 & \cellcolor{hbpHigh!6.57!hbpLow}41.9 \\
      \specialrule{0.30pt}{2.8pt}{2pt}
      \textbf{Median}
        & \textbf{58.2} & \textbf{51.5} & \textbf{60.6} & \textbf{49.9} & \textbf{49.4} & \textbf{53.5} & \textbf{63.8}
        & \textbf{60.5} & \textbf{55.5} & \textbf{48.8} & \textbf{58.7} & \textbf{69.1} & \textbf{47.0} \\
      \specialrule{0.30pt}{2.1pt}{0pt}
    \end{tabular}%
  }
  \par\vspace{1.7mm}
  \noindent\makebox[\linewidth][r]{\fontsize{7}{8.2}\selectfont
    Score\enspace 20\enspace
    \hbpscaletick{11.1825mm}{50}\hbpscaletick{18.6375mm}{70}%
    \hbpswatch{0.00}\hbpswatch{0.51}\hbpswatch{1.02}\hbpswatch{1.53}\hbpswatch{2.04}\hbpswatch{2.55}%
    \hbpswatch{3.06}\hbpswatch{3.57}\hbpswatch{4.09}\hbpswatch{4.60}\hbpswatch{5.11}\hbpswatch{5.62}%
    \hbpswatch{6.13}\hbpswatch{6.64}\hbpswatch{7.15}\hbpswatch{7.66}\hbpswatch{8.17}\hbpswatch{8.68}%
    \hbpswatch{10.40}\hbpswatch{14.15}\hbpswatch{17.89}\hbpswatch{21.64}\hbpswatch{25.38}\hbpswatch{29.13}%
    \hbpswatch{32.87}\hbpswatch{36.62}\hbpswatch{40.36}\hbpswatch{44.11}\hbpswatch{47.85}\hbpswatch{51.60}%
    \hbpswatch{54.67}\hbpswatch{57.33}\hbpswatch{60.00}\hbpswatch{62.67}\hbpswatch{65.33}\hbpswatch{68.00}%
    \hbpswatch{70.67}\hbpswatch{73.33}\hbpswatch{76.00}\hbpswatch{78.67}\hbpswatch{81.33}\hbpswatch{84.00}%
    \hbpswatch{86.67}\hbpswatch{89.33}\hbpswatch{92.00}\hbpswatch{94.67}\hbpswatch{97.33}\hbpswatch{100.00}%
    \enspace 100\strut}
  \par
  \endgroup
\end{table}

\subsection{Performance by task attributes}
\label{sec:item-tag-insights}

\begin{table}[!t]
  \centering
  \captionsetup{position=top,width=\linewidth,font=small,justification=raggedright,singlelinecheck=false,skip=7pt}
\caption{\textbf{HealthBench Professional performance by healthcare workflow.}
Task attribute scores are shown for the top 10 models by overall InclusionMed Index, with medians calculated across these 10 models.
Scores span 0--100; colors use a piecewise-linear scale (29--100).}
  \label{tab:hbp-healthcare-workflows}
  \begingroup
  \sffamily\fontsize{7.75}{9.4}\selectfont\color{black}
  \definecolor{hbpLow}{HTML}{FAFCFF}
  \definecolor{hbpHigh}{HTML}{85ADD7}
  \definecolor{hbpWorkflow}{HTML}{E7F0FA}
  \arrayrulecolor{black}
  \setlength{\tabcolsep}{1.25pt}
  \renewcommand{\arraystretch}{1.42}
  \renewcommand{\tabularxcolumn}[1]{m{#1}}
  \newcommand{\hbphead}[1]{{\fontsize{7.1}{8.5}\selectfont
    \renewcommand{\arraystretch}{0.97}\begin{tabular}[c]{@{}c@{}}#1\end{tabular}}}
  \makeatletter
  \newcommand{\hbpcenterrow}[2][c]{%
    \makebox[\linewidth][#1]{%
      \raisebox{\dimexpr(\ht\@arstrutbox-\dp\@arstrutbox-\height+\depth)/2-0.24pt\relax}[0pt][0pt]{#2}}}
  \def\hbpnumberink#1#2.#3\@nil{%
    \kern-\XeTeXglyphbounds1 \XeTeXcharglyph `#1\relax
    #1#2.#3%
    \kern-\XeTeXglyphbounds3 \XeTeXcharglyph `#3\relax}
  \newcommand{\hbpscore}[2][]{\hbpcenterrow{#1\hbpnumberink#2\@nil}}
  \makeatother
  \newcommand{\hbpswatch}[1]{\textcolor{hbpHigh!#1!hbpLow}{\rule{0.62125mm}{1.38mm}}}
  \newcommand{\hbpscaletick}[2]{%
    \rlap{\hspace{#1}\makebox[0pt][c]{%
      \raisebox{-1.2pt}{\rule{0.24pt}{1.2pt}}}}%
    \rlap{\hspace{#1}\makebox[0pt][c]{%
      \raisebox{-6.5pt}{\fontsize{5.5}{6.5}\selectfont #2}}}}
  \begin{tabularx}{\linewidth}{@{}
    >{\raggedright\arraybackslash}m{33mm}
    *{4}{>{\centering\arraybackslash\columncolor{white}[0.86pt][0.86pt]}X}
    >{\centering\arraybackslash\columncolor{white}[0.86pt][0pt]}X@{}}
    & \multicolumn{5}{>{\columncolor{hbpWorkflow}[0.86pt][0pt]}c@{}}{%
      \rule[-3.9pt]{0pt}{14pt}\textbf{Healthcare workflow}} \\
    \addlinespace[3pt]
    \textbf{Model}
    & \hbphead{Treatment planning\\and implementation}
    & \hbphead{Research\\and teaching}
    & \hbphead{Clinical assessment\\and diagnosis}
    & \hbphead{Healthcare services\\and management}
    & \hbphead{Follow-up\\and monitoring} \\
    \specialrule{0.30pt}{3pt}{2.5pt}
    \hbpcenterrow[l]{Gemini 4 Argon}
      & \cellcolor{hbpHigh!8.58!hbpLow}\hbpscore{53.8}
      & \cellcolor{hbpHigh!10.17!hbpLow}\hbpscore{55.4}
      & \cellcolor{hbpHigh!21.91!hbpLow}\hbpscore{59.4}
      & \cellcolor{hbpHigh!60.21!hbpLow}\hbpscore{74.6}
      & \cellcolor{hbpHigh!28.95!hbpLow}\hbpscore{61.8} \\
    \hbpcenterrow[l]{Claude Opus 5.5}
      & \cellcolor{hbpHigh!18.97!hbpLow}\hbpscore{58.4}
      & \cellcolor{hbpHigh!23.67!hbpLow}\hbpscore{60.0}
      & \cellcolor{hbpHigh!32.17!hbpLow}\hbpscore{62.9}
      & \cellcolor{hbpHigh!75.72!hbpLow}\hbpscore{84.5}
      & \cellcolor{hbpHigh!76.81!hbpLow}\hbpscore{85.2} \\
    \hbpcenterrow[l]{Claude Fable 5.1}
      & \cellcolor{hbpHigh!10.47!hbpLow}\hbpscore{55.5}
      & \cellcolor{hbpHigh!16.04!hbpLow}\hbpscore{57.4}
      & \cellcolor{hbpHigh!26.01!hbpLow}\hbpscore{60.8}
      & \cellcolor{hbpHigh!65.53!hbpLow}\hbpscore{78.0}
      & \cellcolor{hbpHigh!53.16!hbpLow}\hbpscore{70.1} \\
    \hbpcenterrow[l]{GPT-6 Astra}
      & \cellcolor{hbpHigh!33.35!hbpLow}\hbpscore{63.3}
      & \cellcolor{hbpHigh!27.77!hbpLow}\hbpscore{61.4}
      & \cellcolor{hbpHigh!46.25!hbpLow}\hbpscore{67.7}
      & \cellcolor{hbpHigh!86.21!hbpLow}\hbpscore{91.2}
      & \cellcolor{hbpHigh!56.60!hbpLow}\hbpscore{72.3} \\
    \hbpcenterrow[l]{GPT-6.1 Sol}
      & \cellcolor{hbpHigh!29.83!hbpLow}\hbpscore{62.1}
      & \cellcolor{hbpHigh!30.71!hbpLow}\hbpscore{62.4}
      & \cellcolor{hbpHigh!36.57!hbpLow}\hbpscore{64.4}
      & \cellcolor{hbpHigh!81.67!hbpLow}\hbpscore{88.3}
      & \cellcolor{hbpHigh!33.93!hbpLow}\hbpscore{63.5} \\
    \hbpcenterrow[l]{DeepSeek V4.1 Flash}
      & \cellcolor{hbpHigh!3.50!hbpLow}\hbpscore{39.1}
      & \cellcolor{hbpHigh!9.88!hbpLow}\hbpscore{55.3}
      & \cellcolor{hbpHigh!14.28!hbpLow}\hbpscore{56.8}
      & \cellcolor{hbpHigh!55.98!hbpLow}\hbpscore{71.9}
      & \cellcolor{hbpHigh!5.26!hbpLow}\hbpscore{44.2} \\
    \hbpcenterrow[l]{Grok 4.7}
      & \cellcolor{hbpHigh!8.03!hbpLow}\hbpscore{52.2}
      & \cellcolor{hbpHigh!14.28!hbpLow}\hbpscore{56.8}
      & \cellcolor{hbpHigh!16.63!hbpLow}\hbpscore{57.6}
      & \cellcolor{hbpHigh!67.26!hbpLow}\hbpscore{79.1}
      & \cellcolor{hbpHigh!8.90!hbpLow}\hbpscore{54.7} \\
    \hbpcenterrow[l]{Kimi K3.1 Pro}
      & \cellcolor{hbpHigh!7.79!hbpLow}\hbpscore{51.5}
      & \cellcolor{hbpHigh!18.39!hbpLow}\hbpscore{58.2}
      & \cellcolor{hbpHigh!19.85!hbpLow}\hbpscore{58.7}
      & \cellcolor{hbpHigh!71.02!hbpLow}\hbpscore{81.5}
      & \cellcolor{hbpHigh!3.22!hbpLow}\hbpscore{38.3} \\
    \hbpcenterrow[l]{Qwen3.8 Max}
      & \cellcolor{hbpHigh!1.83!hbpLow}\hbpscore{34.3}
      & \cellcolor{hbpHigh!3.39!hbpLow}\hbpscore{38.8}
      & \cellcolor{hbpHigh!6.54!hbpLow}\hbpscore{47.9}
      & \cellcolor{hbpHigh!62.40!hbpLow}\hbpscore{76.0}
      & \cellcolor{hbpHigh!3.29!hbpLow}\hbpscore{38.5} \\
    \hbpcenterrow[l]{Kimi K3}
      & \cellcolor{hbpHigh!4.57!hbpLow}\hbpscore{42.2}
      & \cellcolor{hbpHigh!8.31!hbpLow}\hbpscore{53.0}
      & \cellcolor{hbpHigh!14.57!hbpLow}\hbpscore{56.9}
      & \cellcolor{hbpHigh!45.37!hbpLow}\hbpscore{67.4}
      & \cellcolor{hbpHigh!0.00!hbpLow}\hbpscore{29.0} \\
    \specialrule{0.30pt}{2.8pt}{2pt}
    \hbpcenterrow[l]{\textbf{Median}}
      & \hbpscore[\bfseries]{53.0}
      & \hbpscore[\bfseries]{57.1}
      & \hbpscore[\bfseries]{59.1}
      & \hbpscore[\bfseries]{78.5}
      & \hbpscore[\bfseries]{58.2} \\
    \specialrule{0.30pt}{2.1pt}{0pt}
  \end{tabularx}
  \par\vspace{1.7mm}
  \noindent\makebox[\linewidth][r]{\fontsize{7}{8.2}\selectfont
    Score\enspace 29\enspace
    \hbpscaletick{10.92mm}{55}\hbpscaletick{17.22mm}{70}%
    \hbpswatch{0.00}\hbpswatch{0.52}\hbpswatch{1.05}\hbpswatch{1.57}\hbpswatch{2.09}\hbpswatch{2.61}%
    \hbpswatch{3.14}\hbpswatch{3.66}\hbpswatch{4.18}\hbpswatch{4.71}\hbpswatch{5.23}\hbpswatch{5.75}%
    \hbpswatch{6.27}\hbpswatch{6.80}\hbpswatch{7.32}\hbpswatch{7.84}\hbpswatch{8.37}\hbpswatch{8.89}%
    \hbpswatch{12.50}\hbpswatch{16.93}\hbpswatch{21.36}\hbpswatch{25.79}\hbpswatch{30.22}\hbpswatch{34.65}%
    \hbpswatch{39.08}\hbpswatch{43.51}\hbpswatch{47.94}\hbpswatch{52.38}\hbpswatch{55.03}\hbpswatch{57.40}%
    \hbpswatch{59.77}\hbpswatch{62.13}\hbpswatch{64.50}\hbpswatch{66.87}\hbpswatch{69.23}\hbpswatch{71.60}%
    \hbpswatch{73.97}\hbpswatch{76.33}\hbpswatch{78.70}\hbpswatch{81.07}\hbpswatch{83.43}\hbpswatch{85.80}%
    \hbpswatch{88.17}\hbpswatch{90.53}\hbpswatch{92.90}\hbpswatch{95.27}\hbpswatch{97.63}\hbpswatch{100.00}%
    \enspace 100\strut}
  \par
  \endgroup
\end{table}

Role- and capability-specific scores summarize performance across tasks
with different clinical demands. To examine variation within these broader
profiles, we used task-level annotations from the InclusionMed tagging
scheme (\hyperref[fig:inclusionmed-task-attribute-taxonomy]{Figure~\ref*{fig:inclusionmed-task-attribute-taxonomy}})
to compare performance across task categories within individual benchmarks.
We focused on the ten highest-ranked models in the InclusionMed Index,
examining clinical specialties, medical expertise categories, and
healthcare workflows in HealthBench Professional, alongside workflow
differences in PhysicianBench. Additional analyses by physiological system and rubric assessment target are provided in \hyperref[tab:supp-hbp-physiology-rubric]{Table~\ref*{tab:supp-hbp-physiology-rubric}}.

\paragraph{Clinical specialties: Psychiatry has the lowest median score.}
Among the six named clinical specialties in HealthBench Professional,
obstetrics and gynecology had the highest median score across the ten
models (60.6), while psychiatry had the lowest (49.4), followed by
pediatrics (49.9;
\hyperref[tab:index-profiles]{Table~\ref*{tab:index-profiles}}).
All ten models scored lower in surgical specialties than in internal
medicine, with medians of 51.5 and 58.2, respectively. Psychiatry had
a median close to that of surgical specialties but a wider range
across models. DeepSeek V4.1 Flash, Qwen3.8 Max, and Kimi K3 each
had their two lowest specialty scores in psychiatry and general practice.

\paragraph{Medical expertise: Patient communication scores highest.}
Patient communication was the highest-scoring medical expertise
category for all ten models in HealthBench Professional
(median 69.1;
\hyperref[tab:hbp-medical-expertise]{Table~\ref*{tab:hbp-medical-expertise}}).
Medical understanding had the second-highest median (60.5).
Medical calculation and safety and ethics had the two lowest medians,
at 47.0 and 48.8, respectively. Safety and ethics scores ranged
from 21.0 to 62.5, the widest range across the six categories.

\begin{figure}[!t]
  \centering
  \includegraphics[width=\linewidth]{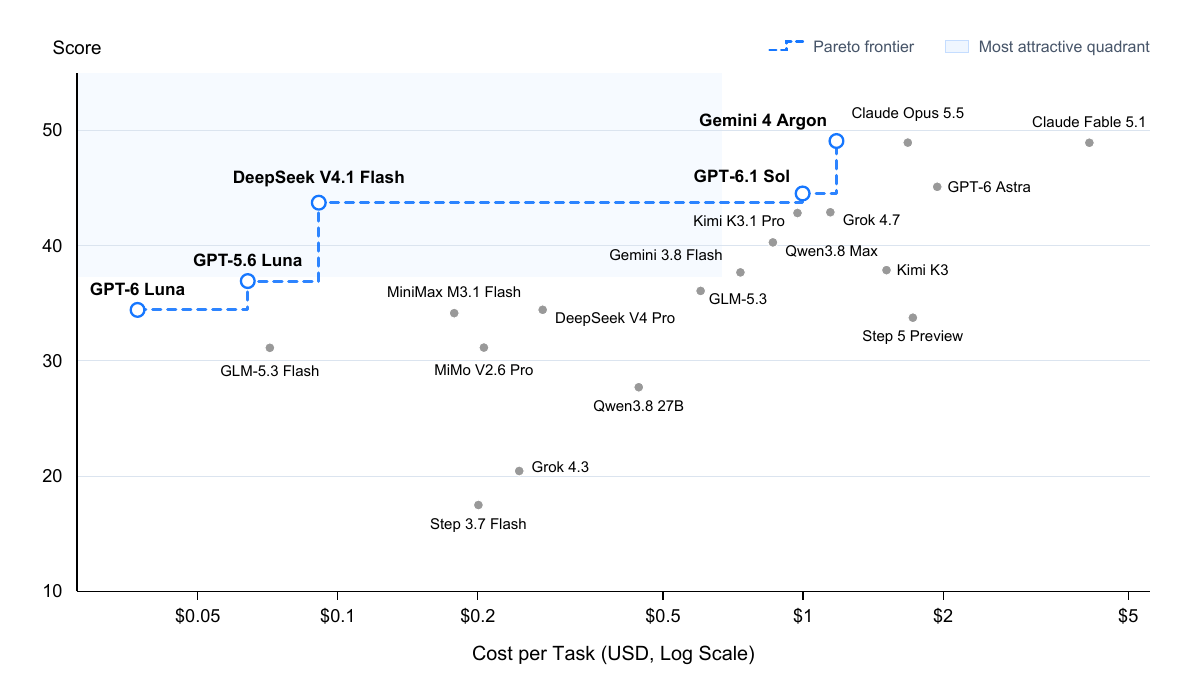}
  \captionsetup{font=small}
\caption{\textbf{InclusionMed Index score versus estimated inference cost.}
Scores and mean costs per task run are averaged across the five benchmarks.
Costs are estimated from recorded token usage and published prices (USD; log scale).
Open circles and the dashed line mark the observed Pareto frontier.}
  \label{fig:quality-resources}
\end{figure}

\paragraph{Healthcare workflows: Treatment tasks have lower median scores.}
All ten models scored lower on treatment planning and implementation
than on clinical assessment and diagnosis in HealthBench Professional
(\hyperref[tab:hbp-healthcare-workflows]{Table~\ref*{tab:hbp-healthcare-workflows}}).
Treatment planning and implementation had the lowest median across
the five workflow categories (53.0). Healthcare services and
management had the highest (78.5), followed by clinical assessment
and diagnosis (59.1).
Eight of the ten models also scored lower on treatment planning
and implementation than on clinical assessment and diagnosis
in PhysicianBench
(\hyperref[tab:supp-physician-workflows]{Table~\ref*{tab:supp-physician-workflows}}).

Checkpoint-level analysis provided further detail on failures in PhysicianBench. In the pooled results from all ten models, clinical reasoning accounted for 469 of the 912 failed checkpoints (51.4\%) and was also the most frequently tested checkpoint category. Median within-category failure rates were highest for action
execution (17.4\%), followed by documentation (14.6\%;
\hyperref[tab:supp-physician-checkpoint-rates]{Table~\ref*{tab:supp-physician-checkpoint-rates}}). These findings highlight the need to assess both clinical reasoning and the ability to reliably carry out clinical actions and document care.

\subsection{Performance--cost trade-offs}
\label{sec:quality-resources}

Inference-time computation can affect reasoning performance~\citep{snell2025testtime}, making inference cost an important consideration alongside performance when selecting models for healthcare tasks. We compare the InclusionMed Index with the equally weighted mean inference cost per task run, calculated as described in \hyperref[sec:inference-cost]{Section~\ref*{sec:inference-cost}}.

Models with similar InclusionMed Index scores differed substantially in estimated cost. Among the highest-scoring models, Claude Fable 5.1 and Claude Opus 5.5 both had scores that rounded to 48.9, yet Fable's estimated cost per task run was approximately 2.5 times that of Opus (\$4.12 versus \$1.68). An even larger difference was observed between Kimi K3 and GPT-5.6 Luna, which scored 37.9 and 36.9, respectively, at estimated costs of \$1.51 and \$0.064 per task run. At these rates, a fixed inference budget could support approximately 24 times as many task runs with GPT-5.6 Luna as with Kimi K3 for the same benchmark mix.

Five models lay on the observed Pareto frontier: GPT-6 Luna, GPT-5.6 Luna, DeepSeek V4.1 Flash, GPT-6.1 Sol, and Gemini 4 Argon. At the lower-cost end, DeepSeek V4.1 Flash achieved a higher Index score than several more expensive models. 

\FloatBarrier

\section{Discussion}
\label{sec:insights}
\label{sec:discussion}
\label{sec:limitations}

The first release of InclusionMed connects benchmark auditing and independent reproduction with model performance and inference cost comparisons, using the Atlas to organize evaluation across healthcare roles and capabilities. This framework makes gaps in benchmark coverage explicit and shows how model rankings vary with the evaluated roles and capabilities, while identifying substantial cost differences between models with similar overall scores. Together, these findings establish a basis for more targeted model evaluation and define the scope of the current evidence. They also identify four priorities for extending InclusionMed: broader representation of healthcare needs, interpretation in the intended context, evaluation under resource constraints, and comparable evidence over time.

\subsection{Evaluation across needs}
\label{app:followup}

The tasks covered by an evaluation determine what it can tell us about model progress. Of the 388 candidate benchmarks we collected, 88\% included English, and most listed an institution in the United States or China first (\hyperref[app:audit-findings]{Appendix~\ref*{app:audit-findings}}). Task-level records also show limitations in coverage: of the 1,034 tasks in the current portfolio, a geographic setting could be determined for 338, of which 318 were associated with the United States (\hyperref[app:outputs]{Appendix~\ref*{app:outputs}}). The current Atlas also includes no benchmark primarily intended to evaluate assistance for patients and the public. Improvements on the current portfolio therefore do not establish whether model performance has improved for these populations and settings.

These gaps give the Atlas concrete directions for expansion. Conversational diagnosis~\citep{tu2025amie} and personal health agents~\citep{heydari2026personalhealthagent} offer examples for developing patient-facing evaluations. Such evaluations could examine how models gather missing information, adapt to changing concerns~\citep{bean2026humaninteraction}, and complete requested actions over sustained interactions. Only 79 candidate benchmarks involved multiple rounds of interaction, leaving a need for further evaluation of these capabilities. Clinicians, patients, and developers from different healthcare settings could help define tasks that reflect local needs and constraints~\citep{alderman2025standing}. We also need to complete existing country and language records and describe care settings in greater detail, so that we can assess whether new tasks genuinely broaden coverage~\citep{qiu2024mmedbench,ariel2026linguistic}. New benchmarks should address underrepresented healthcare needs or reveal capability differences that existing evaluations do not capture~\citep{pfohl2024equitymedqa}. This gives portfolio expansion a purpose beyond increasing the number of benchmarks, while allowing subsequent releases to track both model progress and remaining coverage gaps.

\subsection{Evaluation in context}

An overall ranking reflects performance under a particular combination of benchmarks and weights. The first release assigns equal weights to five benchmarks covering three of the nine Atlas cells; its overall score therefore summarizes performance within this portfolio. Future sensitivity analyses could vary benchmark weights and omit one benchmark at a time to examine which model comparisons remain stable and which depend strongly on the portfolio~\citep{oecd2008composite}.

The relevance of a ranking to a team also depends on how closely the evaluated tasks match the team's needs. The leading model in the current Index differs from the leaders for healthcare professionals, reasoning, and agency (\hyperref[tab:cross-benchmark-atlas-ranks]{Table~\ref*{tab:cross-benchmark-atlas-ranks}}). Even within a benchmark, performance varies across clinical specialties, medical expertise categories, and workflows. The lower scores for healthcare administration professionals and agency in the current portfolio also motivate closer examination of failures on these tasks, although differences in benchmark content and scoring limit direct comparisons across categories. Users can therefore start with the overall ranking and examine more specific performance profiles according to their roles and tasks.

More detailed analysis should help teams decide what to test next. The PhysicianBench results show how separating reasoning, action execution, and documentation reveals different failure patterns within a workflow. Distinguishing activities such as selecting a therapy, adjusting a regimen, entering an order, and documenting care could help identify steps that require closer testing. Where step-level scores are available, linking them to task attributes could help test whether similar failure patterns recur across benchmarks. Reporting the number of tasks or steps in each category would make the basis of these comparisons clearer. This information can help teams narrow their choice of candidate models; suitability for a particular clinical deployment still requires validation in the intended setting~\citep{vasey2022decideai}.

\subsection{Evaluation under constraints}

Cost affects which models a team can choose and how many tasks it can afford to run. In the current benchmark portfolio, Kimi K3 and GPT-5.6 Luna had similar overall scores, but their estimated mean inference costs per task run differed by approximately 24-fold when averaged equally across benchmarks (\hyperref[fig:quality-resources]{Figure~\ref*{fig:quality-resources}}). A large cost difference need not correspond to a similarly large difference in aggregate performance. For users, the next comparison is how much additional performance greater expenditure buys on the tasks they care about~\citep{kapoor2025agents}.

InclusionMed can extend performance--cost comparisons to specific tasks and workflows. A model that is economical for the current benchmark mix may not retain that advantage in tasks requiring prolonged interaction or repeated tool use. Our estimates use token usage and published prices. They exclude failed runs and retries for the four public benchmarks, as well as judge, search, tool, sandbox, and internal platform charges (\hyperref[app:inference-cost]{Appendix~\ref*{app:inference-cost}}), and therefore do not represent total deployment expenditure. Future measurements should cover complete workflows to test whether lower inference costs translate into lower costs in practice. Clinical uses also require a separate assessment of whether these models deliver corresponding clinical benefits.

\subsection{Evaluation over time}

As models improve, tasks that once distinguished their performance may approach their score ceilings~\citep{kiela2021dynabench}. Independent reproduction identified saturation in the tested versions or subsets of three medical knowledge benchmarks and limited discrimination on HealthBench (\hyperref[fig:benchmark-saturation]{Figure~\ref*{fig:benchmark-saturation}}). Continued inclusion in the Index therefore requires periodic reassessment of whether the tasks remain relevant and their scores still distinguish current models. Only 104 of the 388 candidate benchmarks provided sufficient materials for direct reproduction. Clear access conditions, documented scoring procedures, and reproducible execution environments would allow more candidates to be independently evaluated and considered for future releases.

Future releases can track changes in score distributions through periodic reproduction and record the versions of tasks, scoring methods, and execution environments~\citep{gallifant2025tripodllm}. When the portfolio changes, comparisons on shared benchmarks under the same task, scoring, and execution conditions would help distinguish model improvements from changes in evaluation; the coverage gained through new tasks should also be reported separately. Community collaboration can support these efforts by contributing tasks in areas with limited coverage, providing reproducible environments, and examining the evidence behind benchmark inclusion, scoring, and aggregation decisions. As models and healthcare needs change, InclusionMed will also need to explain what each release measures, which results remain comparable, and which judgments still lack sufficient evidence.

\begingroup
\setlength{\parskip}{0.25\baselineskip}
\setlength{\intextsep}{6pt plus 2pt minus 2pt}
\section{Open Collaboration and Contributions}
\label{sec:contributions}

InclusionMed welcomes contributions from anyone interested in medical AI evaluation. Community input can reveal gaps in coverage, improve evaluation resources and help interpret results.

\subsection{Ways to contribute}

\noindent\textbf{Benchmark.} Contributors can share benchmarks, datasets or evaluation tools, describing what they assess and which materials are available. The team works with contributors on review, reproduction and revisions under the assessment principles in \hyperref[sec:scope]{Section~\ref*{sec:scope}}. Accepted benchmarks are released with their scope and limitations.

\noindent\textbf{Benchmark idea.} Contributors can propose a healthcare question for a new benchmark, even without a dataset or evaluation method. The team helps clarify the need and develop an initial task. Any resulting benchmark follows the same review process.

\noindent\textbf{Correction.} Contributors can flag an Index entry and provide evidence for a correction. The team checks the evidence, discusses unresolved questions with the contributor and publishes verified corrections with their sources.

\noindent\textbf{Advisory \& partnership.} Individuals can offer expert guidance, and organizations can propose partnerships. The team and contributors agree on the scope, roles and availability before confirming the collaboration.

\subsection{How to participate}

To participate, visit the
\href{https://inclusionmed.ai}{\textcolor{blue}{\underline{InclusionMed website}}}
and follow the steps below.

\noindent\textbf{Submit a contribution form.} Open the \textbf{Contribute} page and complete the form to describe your contribution, provide contact details and relevant materials, and indicate any support needed.

\noindent\textbf{Develop it together.} The team follows up by email to agree on next steps. Before publication or announcement, the team and contributor confirm the outcome and contribution credits.
Further guidance is available in the
\href{https://github.com/panuvi/inclusionmed/blob/main/CONTRIBUTING.md}{\textcolor{blue}{\underline{GitHub contribution guide}}}.
\par
\endgroup

\clearpage
\appendix
\renewcommand{\theHfigure}{supp.\arabic{figure}}
\renewcommand{\theHtable}{supp.\arabic{table}}
\setcounter{figure}{0}
\renewcommand{\thefigure}{S\arabic{figure}}
\setcounter{table}{0}
\renewcommand{\thetable}{S\arabic{table}}
\section{Supplementary Methods}
\label{app:supplement}
\label{app:methods}

\subsection{Benchmark portfolio}
\label{app:benchmark-portfolio}

\paragraph{Benchmark collection}
We identified candidate benchmarks published between May 2023 and September 2026 from publicly available literature covering topics in medicine, biomedical informatics, and artificial intelligence, as well as institutional releases and community contributions.
For each candidate benchmark, we used LLM-assisted curation to extract information from publications, project websites, and dataset and code repositories and organize it into a structured benchmark card. Each card recorded publication details, evaluation objectives, data sources, and available resources, with links to the original materials. Sources describing the same benchmark were linked to a single record, while benchmark versions and subsets were tracked through version control. This process yielded an initial inventory of 388 candidate benchmarks.

\paragraph{Benchmark auditing}
\label{app:benchmark-auditing}
An agent skill extracted data sources, construction methods, access and licensing conditions, evaluation methods, and reported performance from benchmark cards and linked sources. When task data were accessible, it used LLM APIs to annotate healthcare roles, AI capabilities, and task attributes and summarize their distributions. It also flagged task defects, reference errors, and scoring issues for human review.

The collected information informed the application of the six audit criteria. For the first three criteria, healthcare relevance, trustworthiness, and availability and permitted use, the agent assessed healthcare relevance based on task objectives and intended users; trustworthiness based on data provenance, task construction, annotation procedures, reference standards and scoring methods; and availability and permitted use based on access to data, code and evaluation environments, together with their licensing conditions. The research team spot-checked the agent's exclusion decisions and supporting evidence against the original benchmark materials.
For the remaining three criteria, complementary coverage, timeliness and data independence, and discriminative capacity, the agent provided candidate assessments based on the collected information. At each stage, the research team reviewed each candidate against the applicable criterion and its supporting evidence, deciding whether to exclude it or advance it. Specifically, the agent assessed each criterion as follows. For complementary coverage, it compared candidates with the existing portfolio across healthcare roles and capabilities, as well as task attributes such as language, country and medical expertise. It also examined potential overlap in data sources and task content. For timeliness and data independence, the agent considered publication dates, task sources, changes in relevant guidelines or policies, and safeguards against contamination. For discriminative capacity, it examined score distributions and saturation using reported model results from the same benchmark version. Candidates shortlisted through the audit then proceeded to independent reproduction.

\paragraph{Independent reproduction}
\label{app:benchmark-reproduction}
Following the benchmark audit against six criteria, from healthcare relevance to discriminative capacity, we independently reproduced evaluations of eight shortlisted benchmarks to assess score saturation and discriminative capacity. We evaluated the same eight models across all benchmarks: GPT-6 Astra, Claude Opus 5.5, Gemini 3.8 Flash, DeepSeek V4.1 Flash, Qwen3.8 Max, Qwen3.8 27B, GLM-5.3, and Grok 4.3. The selection covered seven model families and included both open-weight and closed-source models, with flagship and lighter-weight models representing different sizes and capability levels. Including models at different capability levels allowed us to assess saturation among stronger models and discrimination across a broader performance range. Independent reproduction used the same model configurations as the Index evaluation, described in \hyperref[sec:sup_model_list]{Appendix~\ref*{sec:sup_model_list}}. Using scores on a scale from 0 to 100, we applied two exclusion criteria: 1) score saturation, defined as the three highest-scoring models each achieving at least 85 points; and 2) limited discrimination, defined as a sample standard deviation (SD) of no more than 4 points across the eight models. Benchmarks meeting either criterion were excluded.

\subsection{Index protocol}
\label{sec:sup_model_list}

\begin{table}[!htbp]
  \centering
  \captionsetup{
    position=top,
    justification=raggedright,
    singlelinecheck=false,
    skip=5pt
  }
  \caption{\textbf{Model families and releases evaluated in the first InclusionMed Index.}}
  \label{tab:index-model-roster}

  \small
  \setlength{\tabcolsep}{0pt}
  \renewcommand{\arraystretch}{1.16}

  \begin{tabularx}{\linewidth}{@{}p{0.46\linewidth}@{\hspace{1.5em}}>{\raggedright\arraybackslash}X@{}}
    \toprule
    Family & Releases \\
    \midrule
    Claude   & Opus 5.5, Fable 5.1 \\
    DeepSeek & V4.1 Flash 1017, V4 Pro 0813 \\
    Gemini   & 4 Argon, 3.8 Flash \\
    GLM      & 5.3, 5.3 Flash \\
    GPT      & 6 Astra, 6.1 Sol, 6 Luna, 5.6 Luna \\
    Grok     & 4.3, 4.7 \\
    Kimi     & K3, K3.1 Pro \\
    MiniMax  & M3.1 Flash \\
    MiMo     & V2.6 Pro \\
    Qwen     & 3.8 Max 0902, 3.8 27B \\
    Step     & 5 Preview, 3.7 Flash \\
    \bottomrule
  \end{tabularx}
\end{table}

\paragraph{Evaluation protocol}
\label{app:evaluation-protocol}
We evaluated 22 model releases from 11 families.
\hyperref[tab:index-model-roster]{Table~\ref*{tab:index-model-roster}}
lists the evaluated releases, retaining date suffixes where
applicable. These suffixes are omitted from model names in the main
text, figures, and result tables for readability. Specifically,
DeepSeek V4.1 Flash, DeepSeek V4 Pro, and Qwen3.8 Max refer to
DeepSeek V4.1 Flash 1017, DeepSeek V4 Pro 0813, and Qwen3.8 Max 0902,
respectively. All reported results apply to these specific releases.
Responses to HealthBench Professional and InclusionMed-DxFiles cases were
obtained via direct API calls. For PhysicianBench, we used
the benchmark's electronic health record agent in a FHIR environment.
Each model used the same requested reasoning-effort setting across benchmarks: \texttt{xhigh} for the two Qwen releases and \texttt{high} for the other 20 releases. Other generation parameters were not explicitly overridden in the evaluation configuration; their effective values depended on the defaults applied by the serving route and evaluation framework.
For both CHI-Bench and ALE-CLI v1.0: Health and Medicine,
we used Codex for the four GPT releases, Gemini CLI for the two
Gemini releases, and Claude Code for the remaining 16 releases.
The benchmark-specific versions of the agent harnesses are listed in
\hyperref[tab:cli-versions]{Table~\ref*{tab:cli-versions}}.
For HealthBench Professional, API errors, connectivity failures, rate limits and provider safety blocks triggered automatic retries, with a maximum of 10 retries.
For PhysicianBench and CHI-Bench, runs affected by these issues were retried until a valid run was obtained for every sample. For ALE-CLI v1.0: Health and Medicine, API errors, connectivity failures and rate limits were subject to a maximum of three retries. Each ALE-CLI task also had a time limit specified in its task card, typically 7,200\,s; runs exceeding this limit were terminated.

\begin{table}[!htbp]
  \centering
  \captionsetup{
    position=top,
    justification=raggedright,
    singlelinecheck=false,
    skip=5pt
  }
  \caption{\textbf{Harness settings for evaluated models.}
Specific version numbers or release tags of the agent harnesses used for CHI-Bench and
ALE-CLI v1.0: Health and Medicine.}
  \label{tab:cli-versions}

  \small
  \setlength{\tabcolsep}{0pt}
  \renewcommand{\arraystretch}{1.12}

  \begin{tabular*}{\linewidth}{
    @{\extracolsep{\fill}}lllcc@{}
  }
    \toprule
    Benchmark & Model & Harness & Version / Tag of Harness \\
    \midrule

    \multirow[t]{3}{*}{CHI-Bench}
      & GPT series
      & Codex
      & \texttt{0.128.0} \\
      & Gemini series
      & Gemini CLI
      & \texttt{0.67.0} \\
      & Other models
      & Claude Code
      & \texttt{2.1.123} \\

    \midrule

    \multirow[t]{4}{*}{%
       \begin{tabular}[t]{@{}l@{}}
        ALE-CLI v1.0:\\
        Health and\\
        Medicine
      \end{tabular}%
    }
      & GPT series
      & Codex
      & \texttt{0.0.0-agenthle-20260614}  \\
      & Gemini series
      & Gemini CLI
      & \texttt{0.38.8-agenthle}  \\
      & Claude Opus 5.5
      & Claude Code
      & \texttt{2.1.280} \\
      & Other models
      & Claude Code
      & \texttt{2.1.170} \\

    \bottomrule
  \end{tabular*}
\end{table}

\paragraph{Benchmark scoring}
\label{app:benchmark-scoring}
For the four public benchmarks, we extracted the scoring logic from the original implementations and integrated it into our internal evaluation infrastructure. The scoring logic and evaluation rules were preserved without modification. Where LLM judges were used, their model configurations matched those of the official implementations. HealthBench Professional used GPT-5.4 as the judge, with reasoning effort set to \texttt{low}.
After the applicable retry policy, any remaining invalid runs and runs terminated at the task time limit were scored as failures and retained in the benchmark-score denominator. HealthBench Professional samples that remained invalid after the retry limit also counted towards the refusal rate. This operational measure includes unresolved technical failures and provider safety blocks, rather than only explicit refusals in model responses.

\begin{figure}[!htbp]
  \centering
  \includegraphics[width=\linewidth]{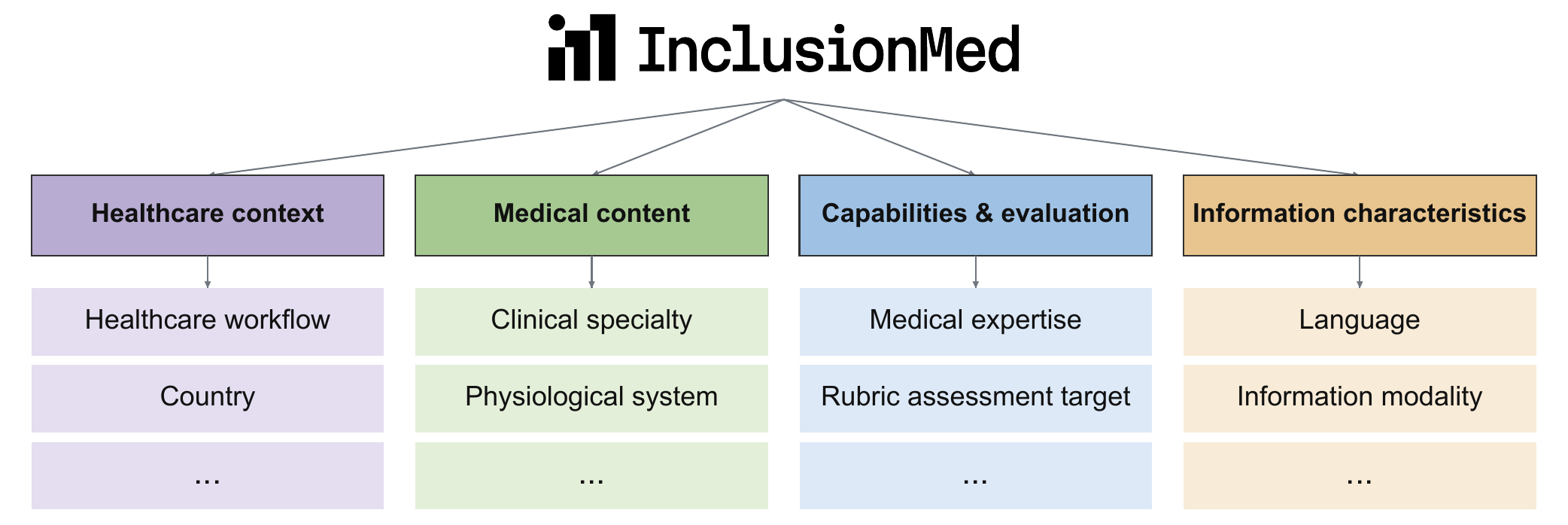}
    \caption{\textbf{InclusionMed tagging scheme.}
    Eight task attributes are grouped into four dimensions: healthcare context, medical content, capabilities and evaluation, and information characteristics.}
  \label{fig:inclusionmed-task-attribute-taxonomy}
\end{figure}

\paragraph{Performance by task attributes}
\label{app:fine-grained-task-characterization}
We characterize tasks using eight attributes grouped into four dimensions (\Cref{fig:inclusionmed-task-attribute-taxonomy}): healthcare context (healthcare workflow and country), medical content (clinical specialty and physiological system), capabilities and evaluation (medical expertise and rubric assessment target), and information characteristics (language and information modality). Each task is assigned one label per attribute, except for rubric assessment target, which is assigned one label per rubric item.

\paragraph{Index aggregation}
\label{app:index-aggregation}
The overall Index and role- and capability-specific scores follow \hyperref[sec:index-aggregation]{Section~\ref*{sec:index-aggregation}}. Task-attribute analyses retain length-adjusted rubric scores for HealthBench Professional (except for rubric assessment targets, for which we average the pass rates of individual rubric items using their absolute weights) and case-level success rates for PhysicianBench. Category medians summarize the scores of the ten highest-ranked models in the overall Index.

\paragraph{Inference cost}
\label{app:inference-cost}

Inference costs mentioned in \hyperref[sec:inference-cost]{Section~\ref*{sec:inference-cost}} were estimated using API token prices published on the official websites of model providers. Table~\ref{tab:model-api-pricing} summarizes the reference prices for the 22 evaluated models as of 13 November 2026, expressed in US dollars per million tokens. Specifically, we applied peak-hour rates to DeepSeek models. For Qwen models, we used the prices listed for the Singapore region and estimated cache-write costs using the rates applicable to implicit caching. Cache-write rates were based on retention periods of 5 minutes for Claude and Kimi K3.1 Pro and 30 minutes for GPT models.

The official pricing sources are as follows:
\par
\begingroup
\footnotesize\raggedright
\begin{itemize}
\setlength{\itemsep}{1pt}
\setlength{\parsep}{0pt}
  \item \textbf{OpenAI}: \url{https://developers.openai.com/api/docs/pricing}

  \item \textbf{Anthropic}: \url{https://platform.claude.com/docs/en/about-claude/pricing}

  \item \textbf{Google}: \url{https://aistudio.google.com/docs/pricing}

  \item \textbf{DeepSeek}: \url{https://api-docs.deepseek.com/quick_start/pricing/}

  \item \textbf{Z.AI}: \url{https://docs.z.ai/guides/overview/pricing}

  \item \textbf{xAI}: \url{https://docs.x.ai/developers/pricing}

  \item \textbf{Moonshot AI (Kimi)}: \url{https://platform.kimi.ai/docs/pricing/chat}

  \item \textbf{Alibaba (Qwen)}: \href{https://docs.qwencloud.com/developer-guides/getting-started/model-prices}{\nolinkurl{https://docs.qwencloud.com/developer-guides/getting-started/}\allowbreak\mbox{\texttt{model-prices}}}

  \item \textbf{MiniMax}: \url{https://platform.minimax.io/docs/pricing/overview}

  \item \textbf{Xiaomi (MiMo)}: \url{https://mimo.mi.com/models/en-US/mimo-v2-6-pro}

  \item \textbf{StepFun}: \url{https://platform.stepfun.ai/docs/en/guides/pricing/details}
\end{itemize}
\par
\endgroup

\begingroup
\setlength{\intextsep}{6pt}
\begin{table}[!htbp]
\centering
\caption{\textbf{Reference API token prices for the evaluated models.} Prices are in US dollars per million tokens, as of 13 November 2026.}
\label{tab:model-api-pricing}

\begingroup
\footnotesize
\setlength{\tabcolsep}{5pt}
\renewcommand{\arraystretch}{1.10}
\begin{tabularx}{\linewidth}{@{}l *{4}{>{\centering\arraybackslash}X}@{}}
\toprule
Model & Input & Cache hit & Output & Cache write \\
\midrule
Claude Opus 5.5     & 4.00  & 0.20   & 20.00 & 5.00  \\
Claude Fable 5.1    & 10.00 & 0.25   & 50.00 & 12.50 \\
DeepSeek V4.1 Flash & 0.30  & 0.006  & 1.20  & ---   \\
DeepSeek V4 Pro     & 1.32  & 0.044  & 3.96  & ---   \\
Gemini 4 Argon      & 2.00  & 0.10   & 10.00 & ---   \\
Gemini 3.8 Flash    & 0.75  & 0.075  & 3.75  & ---   \\
GLM-5.3             & 1.40  & 0.26   & 4.40  & ---   \\
GLM-5.3 Flash       & 0.15  & 0.03   & 0.50  & ---   \\
GPT-6 Astra         & 10.00 & 1.00   & 50.00 & 12.50 \\
GPT-6.1 Sol         & 2.00  & 0.10   & 10.00 & 2.50  \\
GPT-6 Luna          & 0.10  & 0.01   & 0.50  & 0.125 \\
GPT-5.6 Luna        & 0.20  & 0.02   & 1.20  & 0.25  \\
Grok 4.3            & 1.25  & 0.20   & 2.50  & ---   \\
Grok 4.7            & 2.00  & 0.50   & 6.00  & ---   \\
Kimi K3             & 3.00  & 0.30   & 15.00 & 3.00  \\
Kimi K3.1 Pro       & 2.00  & 0.20   & 10.00 & 2.00  \\
MiniMax M3.1 Flash  & 0.15  & 0.015  & 0.60  & ---   \\
MiMo V2.6 Pro       & 0.435 & 0.0036 & 0.87  & ---   \\
Qwen3.8 Max         & 2.00  & 0.25   & 6.00  & 2.00  \\
Qwen3.8 27B         & 0.50  & 0.10   & 3.00  & 0.5 \\
Step 5 Preview      & 1.00  & 0.10   & 2.70  & ---   \\
Step 3.7 Flash      & 0.20  & 0.04   & 1.15  & ---   \\
\bottomrule
\end{tabularx}
\endgroup
\end{table}
\endgroup

\clearpage
\section{Supplementary Results}
\label{app:supplementary_results}

\begin{figure}[!htbp]
    \centering
    \includegraphics[width=\linewidth]{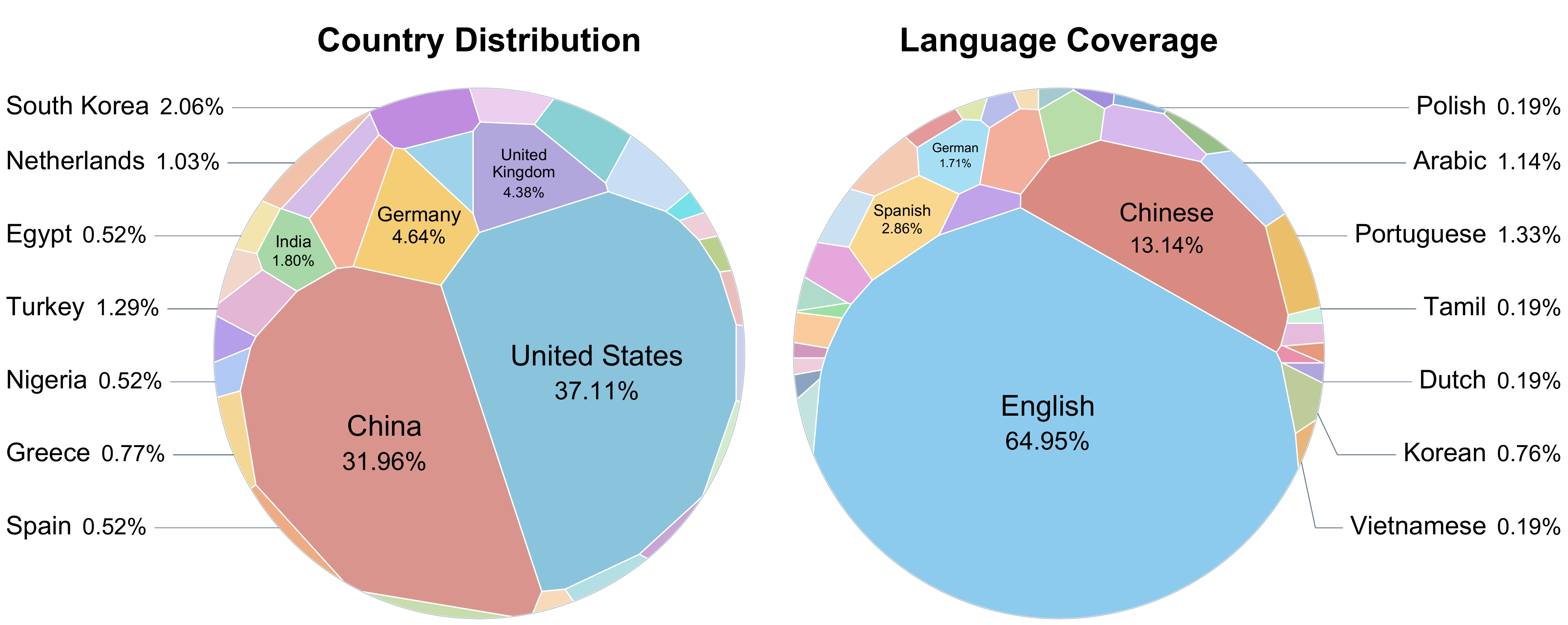}
    \captionsetup{font=small}
    \caption{\textbf{Geographic origins and language coverage of 388 medical benchmarks.} Left: benchmark shares by country of the first listed releasing institution. Right: language shares across 525 reported language assignments, with multilingual benchmarks counted once per reported language.}
    \label{fig:benchmark_distribution}
\end{figure}

\subsection{Benchmark audit results}

\label{app:audit-findings}
\label{app:saturation-exclusions}

\paragraph{The emergence of benchmarks}
We surveyed 388 candidate benchmarks released between May 2023 and September 2026, with 370 released in 2025 or 2026. The inventory was dominated by chat benchmarks but also included 49 agentic benchmarks, an emerging focus of research interest.

\paragraph{Geographic concentration and English predominance}
Geographic origin was determined by the country of the first listed releasing institution. The United States and China accounted for 37.11\% and 31.96\% of benchmarks, respectively. English was represented in 87.89\% of the 388 benchmarks. We counted each benchmark once per reported language, yielding 525 language assignments. English accounted for 64.95\% of these assignments, Chinese for 13.14\%, and Spanish for 2.86\% (\hyperref[fig:benchmark_distribution]{Figure~\ref*{fig:benchmark_distribution}}).

\paragraph{Limited support for direct reproduction}
Only 104 of the 388 candidate benchmarks (26.80\%) provided sufficient materials for direct reproduction. Another 111 benchmarks provided partial materials, while access to materials for 80 benchmarks was restricted or conditional. The remaining 93 benchmarks had no publicly available reproduction materials.

\begin{figure}[!htbp]
  \centering
  \includegraphics[width=\linewidth]{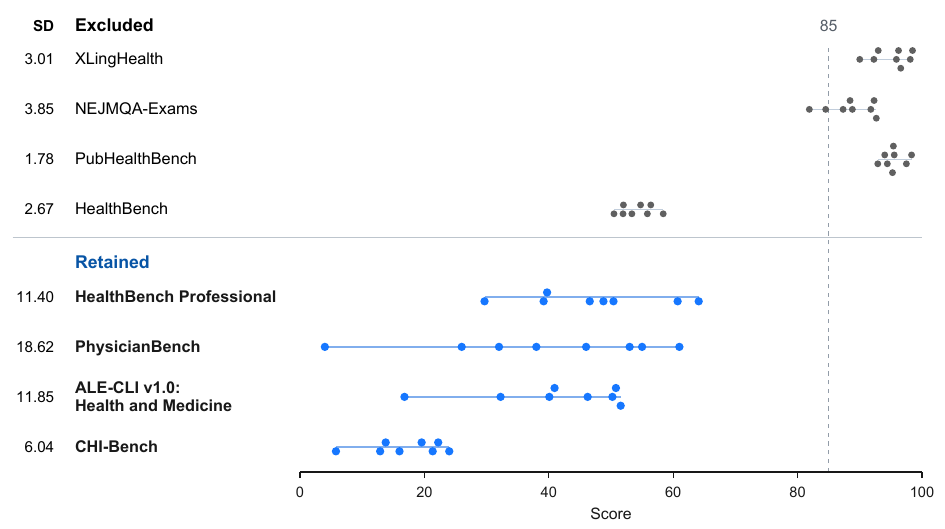}
  \captionsetup{font=small}
    \caption{\textbf{Score saturation and discriminative capacity across eight independently reproduced benchmarks.}
    Points show model scores; sample standard deviations (SDs) are listed on the left.
    Benchmarks were excluded if at least three models scored $\geq 85$ (dashed line) or SD was $\leq 4$.}
  \label{fig:benchmark-saturation}
\end{figure}

\subsection{Index evaluation results}
\label{app:outputs}

In addition to healthcare workflow, clinical specialty, and medical expertise analyzed in the main text, the eight task attributes include country, language, information modality, physiological system, and rubric assessment target. We report the distribution of all eight attributes and, after examining these distributions, further analyze HealthBench Professional performance by physiological system and rubric assessment target.

\subsubsection{Distribution of task attributes}
\paragraph{Clinical assessment and diagnosis led across healthcare workflows, with internal medicine specialties and clinical decision-making most represented.}
Healthcare workflow describes the main healthcare activity required by a task. Clinical assessment and diagnosis was the most common category in the portfolio, accounting for 386 tasks (37.33\%), followed by treatment planning and implementation with 342 (33.08\%), research and teaching with 227 (21.95\%), healthcare services and management with 72 (6.96\%), and follow-up and monitoring with 7 (0.68\%). Clinical specialty groups tasks by the specialty most closely associated with their main professional requirements. Internal medicine specialties accounted for 528 tasks (51.06\%) and surgical specialties for 202 (19.54\%). Pediatrics, obstetrics and gynecology, general practice, and psychiatry together accounted for 207 tasks (20.02\%), with 43--64 tasks (4.16\%--6.19\%) per specialty. Medical expertise describes the principal medical capability required to complete a task. Clinical decision-making was most common, with 632 tasks (61.12\%), followed by medical understanding, safety and ethics, and evidence-based medicine, with 145 (14.02\%), 107 (10.35\%), and 84 (8.12\%), respectively. Medical calculation and patient communication were least common, with 34 tasks (3.29\%) and 32 (3.10\%), respectively.

\paragraph{Identifiable task settings mainly reflect the healthcare context of the United States, and inputs are predominantly English and text-only.}
Country identifies the geographic setting relevant to a benchmark task. Across the five included benchmarks, this attribute could be determined for 338 of the 1,034 tasks (32.69\%), representing 13 geographic settings, with 318 of these tasks (94.08\%) associated with the United States. Language records the primary language of the task input. English accounted for 964 tasks (93.23\%), while the remaining tasks spanned eight non-English languages, most commonly Portuguese (29 tasks), Estonian (21), and Spanish (12). Information modality describes the input a model must interpret to complete a task as instructed. Under this definition, 1,032 tasks (99.81\%) were text-only, while the remaining two (0.19\%), both from ALE-CLI v1.0: Health and Medicine, required visual inspection of images and were therefore classified as text-and-image tasks. Given their highly concentrated distributions, these three attributes were not used to stratify model performance in the analyses below.

\paragraph{Nervous system tasks were most prevalent, and medical accuracy led rubric assessment targets.}
Physiological system classifies tasks by the primary body system involved. Nervous system tasks were most common, accounting for 189 of the 1,034 tasks (18.28\%), followed by cardiovascular and systemic or multisystem tasks with 120 (11.61\%) and 97 (9.38\%), respectively. Each of the remaining 11 categories contained 31--84 tasks (3.00\%--8.12\%). Rubric assessment target describes what each rubric item or checkpoint evaluates. Of the 2,331 items, medical accuracy was most common, with 1,211 items (51.95\%), followed by information completeness and clinical actionability with 312 (13.38\%) and 281 (12.06\%), respectively. Medical safety accounted for 218 items (9.35\%) and information gathering and clarification for 201 (8.62\%). Context and resource adaptation was least common, with 27 items (1.16\%).

\Needspace{6\baselineskip}
\subsubsection{Performance by task attributes}
\paragraph{Models struggle with systemic and multisystem tasks.}
In HealthBench Professional, systemic and multisystem tasks had a median score of 42.2, the lowest among the 14 categories (\hyperref[tab:supp-hbp-physiology-rubric]{Table~\ref*{tab:supp-hbp-physiology-rubric}}). No model reached 60: GPT-6 Astra and GPT-6.1 Sol led with scores of 55.4 and 52.3, respectively. The other eight models scored between 30.7 and 49.0, with Qwen3.8 Max at the lower end of this range.

\paragraph{Information gathering and clarification is the weakest rubric category for most models.}
Among the eight rubric assessment targets, information gathering and clarification had the lowest median weighted pass rate across the ten models (66.1\%), while context and resource adaptation had the highest (79.7\%) (\hyperref[tab:supp-hbp-rubric-cont]{Table~\ref*{tab:supp-hbp-rubric-cont}}). Information gathering and clarification ranked last among the eight targets for seven of the ten models. Across all ten, its weighted pass rate ranged from 55.2\% for DeepSeek V4.1 Flash to 74.2\% for GPT-6 Astra. For context and resource adaptation, weighted pass rates ranged from 69.6\% for Kimi K3 to 85.9\% for GPT-6.1 Sol.

\begin{table}[!htbp]
  \centering
  \captionsetup{position=top,width=\linewidth,font=small,
    justification=raggedright,singlelinecheck=false,skip=7pt}
\caption{\textbf{HealthBench Professional performance by physiological system and rubric assessment target.}
Task attribute scores are shown for the top 10 models by overall InclusionMed Index, with medians calculated across these 10 models.
Physiological scores are length-adjusted; rubric pass rates use absolute criterion weights without length adjustment.}
  \label{tab:supp-hbp-physiology-rubric}
  \label{tab:supp-hbp-rubric-cont}

  \begingroup
  \sffamily\fontsize{7.4}{9.1}\selectfont\color{black}
  \definecolor{hbpSuppLow}{HTML}{FAFCFF}
  \definecolor{hbpSuppHigh}{HTML}{85ADD7}
  \arrayrulecolor{black}
  \setlength{\tabcolsep}{0pt}
  \setlength{\fboxsep}{0pt}
  \renewcommand{\arraystretch}{1}
  \newcommand{\hbpSuppLabelWidth}{108pt}
  \newcommand{\hbpSuppGap}{0.86pt}
  \newcommand{\hbpSuppMedianGap}{2.6pt}
  \newcommand{\hbpSuppWidth}{%
    \dimexpr(\linewidth-\hbpSuppLabelWidth-\hbpSuppGap*10-\hbpSuppMedianGap)/11\relax}
  \newcommand{\hbpSuppRowHeight}{9.6pt}
  \newcommand{\hbpSuppRowDepth}{6.4pt}
  \newcommand{\hbpSuppCenter}[1]{%
    \sbox{0}{#1}%
    \raisebox{\dimexpr(\hbpSuppRowHeight-\hbpSuppRowDepth-\ht0+\dp0)/2\relax}%
      [\hbpSuppRowHeight][\hbpSuppRowDepth]{\usebox{0}}}
  \sbox{0}{68.0}%
  \edef\hbpSuppDataLift{\the\dimexpr
    (\hbpSuppRowHeight-\hbpSuppRowDepth-\ht0+\dp0)/2\relax}
  \newcommand{\hbpSuppDataText}[1]{%
    \raisebox{\hbpSuppDataLift}[\hbpSuppRowHeight][\hbpSuppRowDepth]{#1}}
  \newcommand{\hbpSuppLabel}[1]{%
    \makebox[\hbpSuppLabelWidth][l]{%
      \hbpSuppDataText{\fontsize{7.15}{8.1}\selectfont #1}}}
  \newcommand{\hbpSuppWrappedLabel}[2]{%
    \makebox[\hbpSuppLabelWidth][l]{%
      \hbpSuppCenter{\fontsize{7.15}{7.8}\selectfont
        \vbox{\hbox{#1}\hbox{#2}}}}}
  \newcommand{\hbpSuppHead}[1]{%
    \makebox[\hbpSuppWidth][c]{%
      \hbpSuppCenter{\fontsize{6.25}{7.5}\selectfont
        \renewcommand{\arraystretch}{0.94}%
        \begin{tabular}[c]{@{}c@{}}#1\end{tabular}}}}
  \newcommand{\hbpSuppCell}[2]{%
    \colorbox{hbpSuppHigh!#1!hbpSuppLow}{%
      \makebox[\hbpSuppWidth][c]{\hbpSuppDataText{#2}}}}
  \newcommand{\hbpSuppGroup}[1]{%
    \multicolumn{12}{@{}l@{}}{\makebox[\linewidth][l]{%
      \rule[-4pt]{0pt}{12pt}\textbf{#1}}}}
  \newcommand{\hbpSuppScaleRow}[1]{%
    \multicolumn{12}{@{}r@{}}{#1\kern8pt}}
  \newcommand{\hbpSuppMedian}[1]{%
    \makebox[\hbpSuppWidth][c]{\hbpSuppDataText{\textbf{#1}}}}
  \newcommand{\hbpSuppSwatch}[1]{%
    \textcolor{hbpSuppHigh!#1!hbpSuppLow}{\rule{1.6pt}{2.2pt}}}
  \newcommand{\hbpSuppTick}[2]{%
    \rlap{\kern#1\makebox[0pt][c]{\raisebox{0.7pt}{\rule{0.25pt}{1.3pt}}}}%
    \rlap{\kern#1\makebox[0pt][c]{\raisebox{-4.6pt}{#2}}}}
  \newcommand{\hbpSuppLegendCenter}[1]{%
    \raisebox{\dimexpr(6.2pt-\height+\depth)/2\relax}{#1}}
  \newcommand{\hbpSuppLegend}[5]{%
    \raisebox{1.5pt}[8pt][6pt]{\fontsize{6.3}{7.3}\selectfont
      \hbpSuppLegendCenter{#1}\kern.55em
      \hbpSuppLegendCenter{#2}\kern4pt
      \makebox[76.8pt][l]{%
        \rlap{\raisebox{2pt}{#3}}%
        {\fontsize{5.3}{6.4}\selectfont #4}}%
      \kern4pt\makebox[17pt][l]{\hbpSuppLegendCenter{#5}}}}

  \begin{tabular}{@{}l*{10}{@{\hspace{\hbpSuppGap}}c}@{\hspace{\hbpSuppMedianGap}}c@{}}
    \specialrule{0.30pt}{0pt}{2.5pt}
    \hbpSuppLabel{\textbf{Category}}
    & \hbpSuppHead{Gemini 4\\Argon}
    & \hbpSuppHead{Claude\\Opus 5.5}
    & \hbpSuppHead{Claude\\Fable 5.1}
    & \hbpSuppHead{GPT-6\\Astra}
    & \hbpSuppHead{GPT-6.1\\Sol}
    & \hbpSuppHead{DeepSeek\\V4.1 Flash}
    & \hbpSuppHead{Grok\\4.7}
    & \hbpSuppHead{Kimi K3.1\\Pro}
    & \hbpSuppHead{Qwen3.8\\Max}
    & \hbpSuppHead{Kimi\\K3}
    & \hbpSuppHead{\textbf{Median}} \\
    \specialrule{0.30pt}{2.5pt}{2pt}
    \hbpSuppGroup{Physiological system} \\
    \addlinespace[1.9pt]
    \hbpSuppLabel{Nervous system}
      & \hbpSuppCell{29.92}{58.1} & \hbpSuppCell{50.08}{64.4} & \hbpSuppCell{26.40}{57.0} & \hbpSuppCell{60.96}{67.8} & \hbpSuppCell{61.28}{67.9} & \hbpSuppCell{16.17}{50.9} & \hbpSuppCell{43.36}{62.3} & \hbpSuppCell{33.76}{59.3} & \hbpSuppCell{7.96}{42.1} & \hbpSuppCell{14.31}{48.9} & \hbpSuppMedian{58.7} \\
    \addlinespace[0.72pt]
    \hbpSuppLabel{Sensory system}
      & \hbpSuppCell{40.16}{61.3} & \hbpSuppCell{55.52}{66.1} & \hbpSuppCell{50.40}{64.5} & \hbpSuppCell{36.96}{60.3} & \hbpSuppCell{31.84}{58.7} & \hbpSuppCell{25.12}{56.6} & \hbpSuppCell{20.64}{55.2} & \hbpSuppCell{63.84}{68.7} & \hbpSuppCell{22.24}{55.7} & \hbpSuppCell{30.56}{58.3} & \hbpSuppMedian{59.5} \\
    \addlinespace[0.72pt]
    \hbpSuppLabel{Cardiovascular system}
      & \hbpSuppCell{16.83}{51.6} & \hbpSuppCell{28.00}{57.5} & \hbpSuppCell{32.80}{59.0} & \hbpSuppCell{37.60}{60.5} & \hbpSuppCell{28.64}{57.7} & \hbpSuppCell{13.19}{47.7} & \hbpSuppCell{18.23}{53.1} & \hbpSuppCell{16.83}{51.6} & \hbpSuppCell{5.71}{37.6} & \hbpSuppCell{15.15}{49.8} & \hbpSuppMedian{52.4} \\
    \addlinespace[0.72pt]
    \hbpSuppLabel{Respiratory system}
      & \hbpSuppCell{35.36}{59.8} & \hbpSuppCell{35.36}{59.8} & \hbpSuppCell{41.76}{61.8} & \hbpSuppCell{44.64}{62.7} & \hbpSuppCell{43.04}{62.2} & \hbpSuppCell{17.39}{52.2} & \hbpSuppCell{28.00}{57.5} & \hbpSuppCell{15.15}{49.8} & \hbpSuppCell{7.68}{41.8} & \hbpSuppCell{11.97}{46.4} & \hbpSuppMedian{58.6} \\
    \addlinespace[0.72pt]
    \hbpSuppLabel{Digestive system}
      & \hbpSuppCell{7.12}{41.2} & \hbpSuppCell{16.27}{51.0} & \hbpSuppCell{11.41}{45.8} & \hbpSuppCell{25.12}{56.6} & \hbpSuppCell{31.52}{58.6} & \hbpSuppCell{6.09}{40.1} & \hbpSuppCell{9.08}{43.3} & \hbpSuppCell{13.00}{47.5} & \hbpSuppCell{4.07}{23.9} & \hbpSuppCell{5.68}{37.3} & \hbpSuppMedian{44.5} \\
    \addlinespace[0.72pt]
    \hbpSuppLabel{Urinary system}
      & \hbpSuppCell{82.08}{74.4} & \hbpSuppCell{77.60}{73.0} & \hbpSuppCell{65.76}{69.3} & \hbpSuppCell{76.32}{72.6} & \hbpSuppCell{76.96}{72.8} & \hbpSuppCell{28.32}{57.6} & \hbpSuppCell{31.20}{58.5} & \hbpSuppCell{49.44}{64.2} & \hbpSuppCell{18.04}{52.9} & \hbpSuppCell{36.00}{60.0} & \hbpSuppMedian{66.8} \\
    \addlinespace[0.72pt]
    \hbpSuppLabel{Reproductive system}
      & \hbpSuppCell{28.00}{57.5} & \hbpSuppCell{65.76}{69.3} & \hbpSuppCell{50.72}{64.6} & \hbpSuppCell{76.32}{72.6} & \hbpSuppCell{75.68}{72.4} & \hbpSuppCell{18.97}{53.9} & \hbpSuppCell{35.36}{59.8} & \hbpSuppCell{48.48}{63.9} & \hbpSuppCell{10.20}{44.5} & \hbpSuppCell{17.29}{52.1} & \hbpSuppMedian{61.9} \\
    \addlinespace[0.72pt]
    \hbpSuppLabel{Endocrine system}
      & \hbpSuppCell{18.13}{53.0} & \hbpSuppCell{19.53}{54.5} & \hbpSuppCell{18.23}{53.1} & \hbpSuppCell{36.64}{60.2} & \hbpSuppCell{30.88}{58.4} & \hbpSuppCell{14.40}{49.0} & \hbpSuppCell{18.79}{53.7} & \hbpSuppCell{19.53}{54.5} & \hbpSuppCell{9.73}{44.0} & \hbpSuppCell{19.16}{54.1} & \hbpSuppMedian{53.9} \\
    \addlinespace[0.72pt]
    \hbpSuppLabel{Musculoskeletal system}
      & \hbpSuppCell{36.64}{60.2} & \hbpSuppCell{56.80}{66.5} & \hbpSuppCell{45.28}{62.9} & \hbpSuppCell{51.36}{64.8} & \hbpSuppCell{43.36}{62.3} & \hbpSuppCell{15.80}{50.5} & \hbpSuppCell{23.52}{56.1} & \hbpSuppCell{31.20}{58.5} & \hbpSuppCell{5.59}{36.6} & \hbpSuppCell{16.83}{51.6} & \hbpSuppMedian{59.3} \\
    \addlinespace[0.72pt]
    \hbpSuppLabel{Integumentary system}
      & \hbpSuppCell{22.24}{55.7} & \hbpSuppCell{20.64}{55.2} & \hbpSuppCell{13.84}{48.4} & \hbpSuppCell{45.28}{62.9} & \hbpSuppCell{47.52}{63.6} & \hbpSuppCell{5.95}{39.6} & \hbpSuppCell{26.40}{57.0} & \hbpSuppCell{26.08}{56.9} & \hbpSuppCell{4.98}{31.5} & \hbpSuppCell{10.76}{45.1} & \hbpSuppMedian{55.4} \\
    \addlinespace[0.72pt]
    \hbpSuppLabel{Hematologic system}
      & \hbpSuppCell{19.53}{54.5} & \hbpSuppCell{13.56}{48.1} & \hbpSuppCell{17.57}{52.4} & \hbpSuppCell{56.48}{66.4} & \hbpSuppCell{62.24}{68.2} & \hbpSuppCell{5.47}{35.6} & \hbpSuppCell{14.21}{48.8} & \hbpSuppCell{20.96}{55.3} & \hbpSuppCell{4.81}{30.1} & \hbpSuppCell{6.09}{40.1} & \hbpSuppMedian{50.6} \\
    \addlinespace[0.72pt]
    \hbpSuppLabel{Immune and lymphatic systems}
      & \hbpSuppCell{53.92}{65.6} & \hbpSuppCell{94.56}{78.3} & \hbpSuppCell{58.08}{66.9} & \hbpSuppCell{54.56}{65.8} & \hbpSuppCell{51.68}{64.9} & \hbpSuppCell{4.50}{27.5} & \hbpSuppCell{8.89}{43.1} & \hbpSuppCell{4.45}{27.1} & \hbpSuppCell{3.70}{20.8} & \hbpSuppCell{5.56}{36.3} & \hbpSuppMedian{54.0} \\
    \addlinespace[0.72pt]
    \hbpSuppLabel{Systemic / multisystem}
      & \hbpSuppCell{6.47}{40.5} & \hbpSuppCell{5.92}{39.3} & \hbpSuppCell{7.03}{41.1} & \hbpSuppCell{21.28}{55.4} & \hbpSuppCell{17.48}{52.3} & \hbpSuppCell{9.17}{43.4} & \hbpSuppCell{14.40}{49.0} & \hbpSuppCell{13.56}{48.1} & \hbpSuppCell{4.88}{30.7} & \hbpSuppCell{5.81}{38.4} & \hbpSuppMedian{42.2} \\
    \addlinespace[0.72pt]
    \hbpSuppLabel{Other}
      & \hbpSuppCell{48.16}{63.8} & \hbpSuppCell{67.36}{69.8} & \hbpSuppCell{61.92}{68.1} & \hbpSuppCell{55.52}{66.1} & \hbpSuppCell{55.52}{66.1} & \hbpSuppCell{35.68}{59.9} & \hbpSuppCell{46.24}{63.2} & \hbpSuppCell{49.76}{64.3} & \hbpSuppCell{11.88}{46.3} & \hbpSuppCell{37.60}{60.5} & \hbpSuppMedian{64.0} \\
    \specialrule{0.30pt}{2.5pt}{1.5pt}
    \hbpSuppScaleRow{\raisebox{-1pt}[\height][\depth]{\hbpSuppLegend{Score}{20}{%
      \hbpSuppSwatch{3.60}\hbpSuppSwatch{3.75}\hbpSuppSwatch{3.91}%
      \hbpSuppSwatch{4.06}\hbpSuppSwatch{4.21}\hbpSuppSwatch{4.37}%
      \hbpSuppSwatch{4.52}\hbpSuppSwatch{4.67}\hbpSuppSwatch{4.83}%
      \hbpSuppSwatch{4.98}\hbpSuppSwatch{5.13}\hbpSuppSwatch{5.29}%
      \hbpSuppSwatch{5.44}\hbpSuppSwatch{5.59}\hbpSuppSwatch{5.74}%
      \hbpSuppSwatch{5.90}\hbpSuppSwatch{6.40}\hbpSuppSwatch{7.59}%
      \hbpSuppSwatch{8.78}\hbpSuppSwatch{9.97}\hbpSuppSwatch{11.16}%
      \hbpSuppSwatch{12.35}\hbpSuppSwatch{13.55}\hbpSuppSwatch{14.74}%
      \hbpSuppSwatch{15.93}\hbpSuppSwatch{17.12}\hbpSuppSwatch{18.31}%
      \hbpSuppSwatch{19.50}\hbpSuppSwatch{22.38}\hbpSuppSwatch{26.47}%
      \hbpSuppSwatch{30.55}\hbpSuppSwatch{34.64}\hbpSuppSwatch{38.72}%
      \hbpSuppSwatch{42.81}\hbpSuppSwatch{46.89}\hbpSuppSwatch{50.98}%
      \hbpSuppSwatch{55.06}\hbpSuppSwatch{59.15}\hbpSuppSwatch{63.23}%
      \hbpSuppSwatch{67.32}\hbpSuppSwatch{71.40}\hbpSuppSwatch{75.49}%
      \hbpSuppSwatch{79.57}\hbpSuppSwatch{83.66}\hbpSuppSwatch{87.74}%
      \hbpSuppSwatch{91.83}\hbpSuppSwatch{95.91}\hbpSuppSwatch{100.00}%
    }{%
      \hbpSuppTick{25.6000pt}{40}\hbpSuppTick{44.8000pt}{55}% 
    }{80}}} \\
    \hbpSuppGroup{\raisebox{1.5pt}[\height][\depth]{Rubric assessment target}} \\
    \addlinespace[1.9pt]
    \hbpSuppLabel{Medical accuracy}
      & \hbpSuppCell{18.60}{74.0} & \hbpSuppCell{27.68}{77.4} & \hbpSuppCell{21.28}{75.4} & \hbpSuppCell{21.92}{75.6} & \hbpSuppCell{20.32}{75.1} & \hbpSuppCell{8.10}{66.5}
      & \hbpSuppCell{12.02}{69.3} & \hbpSuppCell{14.26}{70.9} & \hbpSuppCell{6.14}{65.1} & \hbpSuppCell{7.68}{66.2} & \hbpSuppMedian{72.4} \\
    \addlinespace[1.2pt]
    \hbpSuppLabel{Medical safety}
      & \hbpSuppCell{22.56}{75.8} & \hbpSuppCell{39.52}{81.1} & \hbpSuppCell{19.72}{74.8} & \hbpSuppCell{41.44}{81.7} & \hbpSuppCell{38.56}{80.8} & \hbpSuppCell{3.76}{59.4}
      & \hbpSuppCell{13.56}{70.4} & \hbpSuppCell{9.92}{67.8} & \hbpSuppCell{5.92}{64.8} & \hbpSuppCell{4.60}{61.5} & \hbpSuppMedian{72.6} \\
    \addlinespace[1.2pt]
    \hbpSuppLabel{Information completeness}
      & \hbpSuppCell{26.72}{77.1} & \hbpSuppCell{49.44}{84.2} & \hbpSuppCell{40.48}{81.4} & \hbpSuppCell{18.46}{73.9} & \hbpSuppCell{19.16}{74.4} & \hbpSuppCell{18.74}{74.1}
      & \hbpSuppCell{7.82}{66.3} & \hbpSuppCell{18.32}{73.8} & \hbpSuppCell{19.58}{74.7} & \hbpSuppCell{19.44}{74.6} & \hbpSuppMedian{74.5} \\
    \addlinespace[2.3pt]
    \hbpSuppWrappedLabel{Information gathering}{and clarification}
      & \hbpSuppCell{8.94}{67.1} & \hbpSuppCell{15.38}{71.7} & \hbpSuppCell{11.04}{68.6} & \hbpSuppCell{18.88}{74.2} & \hbpSuppCell{13.00}{70.0} & \hbpSuppCell{2.08}{55.2}
      & \hbpSuppCell{3.96}{59.9} & \hbpSuppCell{6.14}{65.1} & \hbpSuppCell{2.92}{57.3} & \hbpSuppCell{4.76}{61.9} & \hbpSuppMedian{66.1} \\
    \addlinespace[2.3pt]
    \hbpSuppLabel{Instruction following}
      & \hbpSuppCell{16.08}{72.2} & \hbpSuppCell{16.36}{72.4} & \hbpSuppCell{18.04}{73.6} & \hbpSuppCell{19.58}{74.7} & \hbpSuppCell{23.84}{76.2} & \hbpSuppCell{14.26}{70.9}
      & \hbpSuppCell{5.88}{64.7} & \hbpSuppCell{11.18}{68.7} & \hbpSuppCell{13.28}{70.2} & \hbpSuppCell{14.12}{70.8} & \hbpSuppMedian{71.6} \\
    \addlinespace[1.2pt]
    \hbpSuppLabel{Clinical actionability}
      & \hbpSuppCell{65.12}{89.1} & \hbpSuppCell{55.84}{86.2} & \hbpSuppCell{52.32}{85.1} & \hbpSuppCell{22.24}{75.7} & \hbpSuppCell{19.02}{74.3} & \hbpSuppCell{30.88}{78.4}
      & \hbpSuppCell{19.72}{74.8} & \hbpSuppCell{27.36}{77.3} & \hbpSuppCell{36.96}{80.3} & \hbpSuppCell{36.32}{80.1} & \hbpSuppMedian{79.3} \\
    \addlinespace[2.3pt]
    \hbpSuppWrappedLabel{Communication and}{interaction}
      & \hbpSuppCell{34.08}{79.4} & \hbpSuppCell{33.12}{79.1} & \hbpSuppCell{38.56}{80.8} & \hbpSuppCell{44.32}{82.6} & \hbpSuppCell{44.32}{82.6} & \hbpSuppCell{2.04}{55.1}
      & \hbpSuppCell{14.68}{71.2} & \hbpSuppCell{26.40}{77.0} & \hbpSuppCell{3.64}{59.1} & \hbpSuppCell{25.12}{76.6} & \hbpSuppMedian{78.0} \\
    \addlinespace[2.8pt]
    \hbpSuppWrappedLabel{Context and resource}{adaptation}
      & \hbpSuppCell{39.20}{81.0} & \hbpSuppCell{39.20}{81.0} & \hbpSuppCell{27.04}{77.2} & \hbpSuppCell{54.24}{85.7} & \hbpSuppCell{54.88}{85.9} & \hbpSuppCell{16.92}{72.8}
      & \hbpSuppCell{16.64}{72.6} & \hbpSuppCell{30.88}{78.4} & \hbpSuppCell{39.84}{81.2} & \hbpSuppCell{12.44}{69.6} & \hbpSuppMedian{79.7} \\
    \specialrule{0.30pt}{2.5pt}{3pt}
    \hbpSuppScaleRow{\hbpSuppLegend{Pass rate (\%)}{50}{%
      \hbpSuppSwatch{0.00}\hbpSuppSwatch{0.43}\hbpSuppSwatch{0.85}%
      \hbpSuppSwatch{1.28}\hbpSuppSwatch{1.70}\hbpSuppSwatch{2.13}%
      \hbpSuppSwatch{2.55}\hbpSuppSwatch{2.98}\hbpSuppSwatch{3.40}%
      \hbpSuppSwatch{3.83}\hbpSuppSwatch{4.26}\hbpSuppSwatch{4.68}%
      \hbpSuppSwatch{5.11}\hbpSuppSwatch{5.53}\hbpSuppSwatch{5.96}%
      \hbpSuppSwatch{7.34}\hbpSuppSwatch{8.83}\hbpSuppSwatch{10.32}%
      \hbpSuppSwatch{11.81}\hbpSuppSwatch{13.30}\hbpSuppSwatch{14.79}%
      \hbpSuppSwatch{16.28}\hbpSuppSwatch{17.77}\hbpSuppSwatch{19.26}%
      \hbpSuppSwatch{21.70}\hbpSuppSwatch{25.11}\hbpSuppSwatch{28.51}%
      \hbpSuppSwatch{31.91}\hbpSuppSwatch{35.32}\hbpSuppSwatch{38.72}%
      \hbpSuppSwatch{42.13}\hbpSuppSwatch{45.53}\hbpSuppSwatch{48.94}%
      \hbpSuppSwatch{52.34}\hbpSuppSwatch{55.74}\hbpSuppSwatch{59.15}%
      \hbpSuppSwatch{62.55}\hbpSuppSwatch{65.96}\hbpSuppSwatch{69.36}%
      \hbpSuppSwatch{72.77}\hbpSuppSwatch{76.17}\hbpSuppSwatch{79.57}%
      \hbpSuppSwatch{82.98}\hbpSuppSwatch{86.38}\hbpSuppSwatch{89.79}%
      \hbpSuppSwatch{93.19}\hbpSuppSwatch{96.60}\hbpSuppSwatch{100.00}%
    }{%
      \hbpSuppTick{23.0400pt}{65}\hbpSuppTick{38.4000pt}{75}% 
    }{100}} \\
  \end{tabular}
  \par
  \endgroup
\end{table}

\begingroup
\definecolor{physBenchHighest}{HTML}{DDEBF9}
\newcommand{\physBenchSwatch}{\textcolor{physBenchHighest}{\rule[0.1ex]{1.2em}{0.75ex}}}
\newcommand{\physBenchHighlight}[1]{%
  \begingroup
  \setlength{\fboxsep}{0pt}%
  \colorbox{physBenchHighest}{\makebox[31pt][c]{\rule[-2.5pt]{0pt}{10pt}#1}}%
  \endgroup}
\newcommand{\physBenchHeader}[1]{%
  \sbox{0}{\textbf{#1}}%
  \raisebox{\dimexpr(\dp0-\ht0)/2\relax}[12pt][12pt]{\usebox{0}}}
\begin{table}[!htbp]
  \centering
  \captionsetup{position=top,width=\linewidth,font=small,
    justification=raggedright,singlelinecheck=false,skip=7pt}
    \caption{\textbf{PhysicianBench performance by healthcare workflow.}
    Case pass rates (0--100) are shown for the top 10 models by overall InclusionMed Index, with medians calculated across these 10 models.
    Follow-up and monitoring ($n=1$) is not shown separately.
    Blue shading highlights the highest score in each column.}
  \label{tab:supp-physician-workflows}
  \sffamily\fontsize{9}{11}\selectfont
  \setlength{\tabcolsep}{3.5pt}
  \renewcommand{\arraystretch}{1.23}
  \renewcommand{\tabularxcolumn}[1]{m{#1}}
  \begin{tabularx}{\linewidth}{@{}
    >{\raggedright\arraybackslash}m{39mm}
    *{3}{>{\centering\arraybackslash}X}@{}}
    \specialrule{0.45pt}{0pt}{2pt}
    \physBenchHeader{Model}
      & \physBenchHeader{\shortstack[c]{\strut Treatment planning\\\strut and implementation}}
      & \physBenchHeader{\shortstack[c]{\strut Clinical assessment\\\strut and diagnosis}}
      & \physBenchHeader{Overall score} \\
    \specialrule{0.30pt}{3pt}{2pt}
    Gemini 4 Argon & 56.5 & \physBenchHighlight{85.7} & 59.0 \\
    Claude Opus 5.5 & \physBenchHighlight{60.9} & 57.1 & 61.0 \\
    Claude Fable 5.1 & \physBenchHighlight{60.9} & 71.4 & \physBenchHighlight{62.0} \\
    GPT-6 Astra & 44.6 & 57.1 & 46.0 \\
    GPT-6.1 Sol & 39.1 & 42.9 & 40.0 \\
    DeepSeek V4.1 Flash & 53.3 & 71.4 & 55.0 \\
    Grok 4.7 & 46.7 & 57.1 & 47.0 \\
    Kimi K3.1 Pro & 41.3 & 28.6 & 41.0 \\
    Qwen3.8 Max & 53.3 & 57.1 & 53.0 \\
    Kimi K3 & 30.4 & 71.4 & 33.0 \\
    \specialrule{0.30pt}{2pt}{2pt}
    \textbf{Median} & \textbf{50.0} & \textbf{57.1} & \textbf{50.0} \\
    \specialrule{0.45pt}{2pt}{0pt}
  \end{tabularx}
\end{table}
\endgroup

\begingroup
\definecolor{physBenchHighest}{HTML}{DDEBF9}
\newcommand{\physBenchSwatch}{\textcolor{physBenchHighest}{\rule[0.1ex]{1.2em}{0.75ex}}}
\newcommand{\physBenchHighlight}[1]{%
  \begingroup
  \setlength{\fboxsep}{0pt}%
  \colorbox{physBenchHighest}{\makebox[31pt][c]{\rule[-2.5pt]{0pt}{10pt}#1}}%
  \endgroup}

\begin{table}[!htbp]
  \centering
  \captionsetup{position=top,width=\linewidth,font=small,
    justification=raggedright,singlelinecheck=false,skip=7pt}
  \caption{\textbf{PhysicianBench failure rates by checkpoint category.}
    Rates are calculated within each checkpoint category across 100 cases per model
    for the top 10 models by overall InclusionMed Index.
    Medians are calculated across these 10 models.
    Blue shading highlights the highest rate in each category.
    Bold in each model row highlights the highest rate for that model.}
  \label{tab:supp-physician-checkpoint-rates}
  \sffamily\fontsize{9}{11}\selectfont
  \setlength{\tabcolsep}{3.5pt}
  \renewcommand{\arraystretch}{1.23}
  \renewcommand{\tabularxcolumn}[1]{m{#1}}
  \begin{tabularx}{\linewidth}{@{}
    >{\raggedright\arraybackslash}m{33mm}
    *{4}{>{\centering\arraybackslash}X}@{}}
    \specialrule{0.45pt}{0pt}{2pt}
    \textbf{Model}
      & \textbf{Data retrieval}
      & \textbf{Clinical reasoning}
      & \textbf{Action execution}
      & \textbf{Documentation} \\[-1.5pt]
    & n = 103 & n = 356 & n = 112 & n = 99 \\
    \specialrule{0.30pt}{3pt}{2pt}
    Gemini 4 Argon & 1.9\% & 10.1\% & 13.4\% & \textbf{15.2\%} \\  % Failed: 2, 36, 15, 15
    Claude Opus 5.5 & 4.9\% & 12.6\% & \textbf{13.4\%} & 10.1\% \\  % Failed: 5, 45, 15, 10
    Claude Fable 5.1 & 2.9\% & 9.6\% & 10.7\% & \textbf{13.1\%} \\  % Failed: 3, 34, 12, 13
    GPT-6 Astra & 1.9\% & 14.9\% & \textbf{19.6\%} & 11.1\% \\  % Failed: 2, 53, 22, 11
    GPT-6.1 Sol & 3.9\% & 14.9\% & \textbf{19.6\%} & 13.1\% \\  % Failed: 4, 53, 22, 13
    DeepSeek V4.1 Flash & 4.9\% & 11.2\% & 12.5\% & \textbf{22.2\%} \\  % Failed: 5, 40, 14, 22
    Grok 4.7 & 7.8\% & 12.9\% & \physBenchHighlight{\textbf{26.8\%}} & 19.2\% \\  % Failed: 8, 46, 30, 19
    Kimi K3.1 Pro & 11.7\% & \physBenchHighlight{19.1\%}
      & \textbf{25.9\%} & 23.2\% \\  % Failed: 12, 68, 29, 23
    Qwen3.8 Max & 7.8\% & 11.2\% & \textbf{15.2\%} & 14.1\% \\  % Failed: 8, 40, 17, 14
    Kimi K3 & \physBenchHighlight{18.4\%} & 15.2\%
      & 22.3\% & \physBenchHighlight{\textbf{34.3\%}} \\  % Failed: 19, 54, 25, 34
    \specialrule{0.30pt}{2pt}{2pt}
    \textbf{Median}
      & \textbf{4.9\%}
      & \textbf{12.8\%}
      & \textbf{17.4\%}
      & \textbf{14.6\%} \\
    \specialrule{0.45pt}{2pt}{0pt}
  \end{tabularx}
\end{table}
\endgroup

\clearpage
\section{InclusionMed-DxFiles}
\label{app:dxfiles}

\subsection{Benchmark design}

InclusionMed-DxFiles assesses diagnostic reasoning on 312 clinical cases adapted from the \emph{Case Records of the Massachusetts General Hospital} in the \emph{New England Journal of Medicine}. The cases span multiple specialties and present complex diagnostic problems. Each task asks a model to interpret the available clinical information and provide a diagnostic analysis and conclusion in a single response. The evaluation focuses on the primary diagnosis and the support for additional diagnostic claims.

\subsection{Case construction}

Cases were selected from a collection of published reports and adapted into a standard clinical format covering the presenting complaint, medical history, examination findings, investigations, and clinical course. The published final diagnosis, pathology findings, and expert discussion were excluded from model inputs. During preparation, the case information was checked against the source reports, and wording that explicitly disclosed the diagnosis was removed. Reference diagnoses were derived from the final diagnoses reported in the source articles. Each reference identified the primary disease explaining the presentation, with associated complications and comorbidities recorded where applicable. Relationships among these conditions were determined from the diagnostic discussion in the source report.

\subsection{Evaluation and scoring}

Responses were evaluated using GPT-6.1 Sol through comparison with the reference diagnoses.
A response received a \emph{perfect} grade when its primary diagnosis matched the reference root cause and all additional diagnoses were supported by the case. The benchmark score is the percentage of the 312 responses graded \emph{perfect}. 
\clearpage
\section{The InclusionMed Collaboration}
\label{app:authors}

{\small\itshape Names within each group are listed alphabetically by first name.\par}

\par\medskip
\noindent\textbf{Project Team}\par\smallskip
\noindent Guangkun Nie\textsuperscript{1}, Huanze Tang\textsuperscript{1}, Juncheng Yang\textsuperscript{1}, Sihan Wang\textsuperscript{1}, Tianbin Li\textsuperscript{1}, Xinyan Deng\textsuperscript{1}, Xinyuan Zhu\textsuperscript{1}, Yifan Zhou\textsuperscript{1}, Yuhao Shen\textsuperscript{1}.\par
\vspace{6pt}

\par\medskip
\noindent\textbf{Collaborators}\par\smallskip
\noindent An Sui\textsuperscript{14},
Bao Li\textsuperscript{3},
Bo Wang\textsuperscript{15},
Changkai Ji\textsuperscript{14},
Cheng Wei\textsuperscript{24},
Chenglong Ma\textsuperscript{14},
Dan Lin\textsuperscript{5},
Dan Yang\textsuperscript{24},
Danfeng Yu\textsuperscript{25},
Fanrui Zhang\textsuperscript{27},
Feng Qiu\textsuperscript{24},
Fuli Feng\textsuperscript{26},
Haoyang Li\textsuperscript{21},
Hongcheng Guo\textsuperscript{14},
Jian Wang\textsuperscript{24},
Jiawei Chen\textsuperscript{16},
Jin Qin\textsuperscript{24},
Jinjie Gu\textsuperscript{24},
Jiyao Liu\textsuperscript{14},
Junyang Wu\textsuperscript{15},
Kevin Lin\textsuperscript{9},
Lang Cao\textsuperscript{8},
Lin Zhang\textsuperscript{24},
Meng Zhou\textsuperscript{24},
Min Cen\textsuperscript{28},
Mingjie Li\textsuperscript{15},
Mingke Yan\textsuperscript{13},
Peng Yi\textsuperscript{24},
Qingqiu Li\textsuperscript{14},
Qingsong Yao\textsuperscript{19},
Rongbin Li\textsuperscript{18},
Shenda Hong\textsuperscript{13},
Shihao Liang\textsuperscript{24},
Shiyi Du\textsuperscript{7},
Shuaiqi Li\textsuperscript{24},
Shuaishuai Li\textsuperscript{24},
Sihan Ding\textsuperscript{11,16},
Sijia Peng\textsuperscript{20},
Tingji Zhou\textsuperscript{24},
Tom Yuan\textsuperscript{2},
Wenhan Wang\textsuperscript{27},
Wenhui Lei\textsuperscript{4},
Xiaohan Ren\textsuperscript{29},
Xiaoyan Yang\textsuperscript{24},
Xiaoyan Zhao\textsuperscript{10},
Xuanzhang Liu\textsuperscript{24},
Yalun Zhu\textsuperscript{24},
Ying Chen\textsuperscript{17},
Yongchao Liu\textsuperscript{24},
Yu Lin\textsuperscript{24},
Yucheng Xing\textsuperscript{10},
Yue Shen\textsuperscript{24},
Yujie Xiao\textsuperscript{13},
Yuliang Du\textsuperscript{24},
Yumou Zhou\textsuperscript{15},
Yutong Xie\textsuperscript{23},
Zhan Qiu\textsuperscript{24},
Zhe Xu\textsuperscript{6},
Zheyao Gao\textsuperscript{12},
Zhiling Ye\textsuperscript{24},
Zilin Lu\textsuperscript{24}.\par
\vspace{6pt}

\par\medskip
\noindent\textbf{Advisory Committee}\par\smallskip
\noindent Benyou Wang\textsuperscript{22}, Yu Qiao\textsuperscript{28}, Zhongyu Wei\textsuperscript{14}, Zuyi Zhu\textsuperscript{24}.\par

\par\medskip
\noindent\textbf{Corresponding Authors}\par\smallskip
\noindent Junjun He (\href{mailto:hejunjun@sii.edu.cn}{\nolinkurl{hejunjun@sii.edu.cn}}); Yuanfeng Ji (\href{mailto:yuanfengji@inclusionmed.ai}{\nolinkurl{yuanfengji@inclusionmed.ai}}).\par

\par\medskip
\noindent\textbf{Affiliations}\par\smallskip
\noindent\begin{minipage}[t]{0.46\linewidth}
\small\raggedright\setlength{\parskip}{4pt}
\noindent\hangindent=1.5em\hangafter=1\makebox[1.5em][l]{\textsuperscript{1}}InclusionMed.\par
\noindent\hangindent=1.5em\hangafter=1\makebox[1.5em][l]{\textsuperscript{2}}Stanford University.\par
\noindent\hangindent=1.5em\hangafter=1\makebox[1.5em][l]{\textsuperscript{3}}Harvard University.\par
\noindent\hangindent=1.5em\hangafter=1\makebox[1.5em][l]{\textsuperscript{4}}University of Pennsylvania.\par
\noindent\hangindent=1.5em\hangafter=1\makebox[1.5em][l]{\textsuperscript{5}}Cornell University.\par
\noindent\hangindent=1.5em\hangafter=1\makebox[1.5em][l]{\textsuperscript{6}}Columbia University.\par
\noindent\hangindent=1.5em\hangafter=1\makebox[1.5em][l]{\textsuperscript{7}}Carnegie Mellon University.\par
\noindent\hangindent=1.5em\hangafter=1\makebox[1.5em][l]{\textsuperscript{8}}University of Illinois Urbana-Champaign.\par
\noindent\hangindent=1.5em\hangafter=1\makebox[1.5em][l]{\textsuperscript{9}}University of Oxford.\par
\noindent\hangindent=1.5em\hangafter=1\makebox[1.5em][l]{\textsuperscript{10}}National University of Singapore.\par
\noindent\hangindent=1.5em\hangafter=1\makebox[1.5em][l]{\textsuperscript{11}}The University of Hong Kong.\par
\noindent\hangindent=1.5em\hangafter=1\makebox[1.5em][l]{\textsuperscript{12}}The Chinese University of Hong Kong.\par
\noindent\hangindent=1.5em\hangafter=1\makebox[1.5em][l]{\textsuperscript{13}}Peking University.\par
\noindent\hangindent=1.5em\hangafter=1\makebox[1.5em][l]{\textsuperscript{14}}Fudan University.\par
\noindent\hangindent=1.5em\hangafter=1\makebox[1.5em][l]{\textsuperscript{15}}Shanghai Jiao Tong University.\par
\noindent\hangindent=1.5em\hangafter=1\makebox[1.5em][l]{\textsuperscript{16}}Zhejiang University.\par
\end{minipage}\hfill%
\begin{minipage}[t]{0.52\linewidth}
\small\raggedright\setlength{\parskip}{4pt}
\noindent\hangindent=1.5em\hangafter=1\makebox[1.5em][l]{\textsuperscript{17}}Xiamen University.\par
\noindent\hangindent=1.5em\hangafter=1\makebox[1.5em][l]{\textsuperscript{18}}Shenzhen University.\par
\noindent\hangindent=1.5em\hangafter=1\makebox[1.5em][l]{\textsuperscript{19}}Renmin University of China.\par
\noindent\hangindent=1.5em\hangafter=1\makebox[1.5em][l]{\textsuperscript{20}}East China University of Science and Technology.\par
\noindent\hangindent=1.5em\hangafter=1\makebox[1.5em][l]{\textsuperscript{21}}Westlake University.\par
\noindent\hangindent=1.5em\hangafter=1\makebox[1.5em][l]{\textsuperscript{22}}The Chinese University of Hong Kong, Shenzhen.\par
\noindent\hangindent=1.5em\hangafter=1\makebox[1.5em][l]{\textsuperscript{23}}Mohamed bin Zayed University of Artificial Intelligence.\par
\noindent\hangindent=1.5em\hangafter=1\makebox[1.5em][l]{\textsuperscript{24}}Ant Group.\par
\noindent\hangindent=1.5em\hangafter=1\makebox[1.5em][l]{\textsuperscript{25}}Chinese Academy of Medical Sciences \& Peking Union Medical College.\par
\noindent\hangindent=1.5em\hangafter=1\makebox[1.5em][l]{\textsuperscript{26}}Anhui Provincial Hospital.\par
\noindent\hangindent=1.5em\hangafter=1\makebox[1.5em][l]{\textsuperscript{27}}Shanghai Innovation Institute.\par
\noindent\hangindent=1.5em\hangafter=1\makebox[1.5em][l]{\textsuperscript{28}}Shanghai Artificial Intelligence Laboratory.\par
\noindent\hangindent=1.5em\hangafter=1\makebox[1.5em][l]{\textsuperscript{29}}Zhongguancun Academy.\par
\end{minipage}\par

\clearpage
\phantomsection
\addcontentsline{toc}{section}{References}
\begingroup
\interlinepenalty=10000
\small\bibliographystyle{inclusionmed-unsrtnat}
\bibliography{Resource/References/ref}
\endgroup

\end{document}